\pdfoutput=1

\documentclass[11pt]{article}

\usepackage[final]{acl}

\usepackage{times}
\usepackage{latexsym}

\usepackage[T1]{fontenc}

\usepackage[utf8]{inputenc}

\usepackage{microtype}

\usepackage{inconsolata}

\usepackage{graphicx}

\usepackage{booktabs,subcaption,amsfonts,dcolumn}
\usepackage{enumitem}

\usepackage{algorithm}
\usepackage{algpseudocode}
\usepackage{multirow}

\usepackage{silence}
\usepackage{silence}
\usepackage{todonotes}
\usepackage{xcolor}
\usepackage{xargs}

\definecolor{red}{RGB}{141,45,57}
\definecolor{dark}{RGB}{55,65,74}
\definecolor{blue}{RGB}{0,105,170}
\definecolor{gold}{RGB}{174,159,109}
\definecolor{gray}{RGB}{175,179,183}
\definecolor{darkgreen}{RGB}{50,110,30}

\newcommandx{\telmo}[2][1=]{\vspace{0.2cm} \todo[linecolor=blue,backgroundcolor=blue!15,bordercolor=blue, #1]{\textbf{Telmo:} #2}}
\newcommandx{\beatriz}[2][1=]{\vspace{0.2cm} \todo[linecolor=darkgreen,backgroundcolor=darkgreen!25,bordercolor=darkgreen, #1]{\textbf{Beatriz:} #2}}
\newcommandx{\andre}[2][1=]{\vspace{0.2cm}\todo[linecolor=gold,backgroundcolor=red!15,bordercolor=gold, #1]{\textbf{André:} #2}}

\title{LegalBench.PT: A Benchmark for Portuguese Law}

\author{Beatriz Canaverde \\
  Instituto Superior Técnico \\
  \texttt{beatriz.canaverde@tecnico.ulisboa.pt} \\\And
  Telmo Pessoa Pires \\
  Equall \\
  \texttt{telmo@equall.com} \\\AND
  Leonor Melo Ribeiro \\
  Georgetown University Law Center \\
  \texttt{leonormeloribeiro@outlook.com} \\\And
  André F. T. Martins \\
  Instituto Superior Técnico \\
  \texttt{andre.t.martins@tecnico.ulisboa.pt} \\}

\begin{document}
\maketitle
\begin{abstract}

The recent application of LLMs to the legal field has spurred the creation of benchmarks across various jurisdictions and languages. However, no benchmark has yet been specifically designed for the Portuguese legal system. In this work, we present LegalBench.PT, the first comprehensive legal benchmark covering key areas of Portuguese law. To develop LegalBench.PT, we first collect long-form questions and answers from real law exams, and then use GPT-4o to convert them into multiple-choice, true/false, and matching formats. Once generated, the questions are filtered and processed to improve the quality of the dataset. To ensure accuracy and relevance, we validate our approach by having a legal professional review a sample of the generated questions. Although the questions are synthetically generated, we show that their basis in human-created exams and our rigorous filtering and processing methods applied result in a reliable benchmark for assessing LLMs' legal knowledge and reasoning abilities. Finally, we evaluate the performance of leading LLMs on LegalBench.PT and investigate potential biases in GPT-4o’s responses. We also assess the performance of Portuguese lawyers on a sample of questions to establish a baseline for model comparison and validate the benchmark.

\end{abstract}

\section{Introduction}

Large Language Models (LLMs) have shown impressive capabilities \citep{claude,jiang2024mixtralexperts,openai2024gpt4technicalreport,dubey2024llama3herdmodels}, driving interest in their legal applications to improve the efficiency and accessibility of legal services. Research has focused on developing legal-specific LLMs \citep{colombo2024saullm7bpioneeringlargelanguage,junior2024jurulegalbrazilianlarge,zhou2024lawgptchineselegalknowledgeenhanced,colombo2024saullm54bsaullm141bscaling}, curating training datasets \citep{NEURIPS2022_bc218a0c,niklaus-etal-2024-multilegalpile,niklaus2024flawnt5empiricalexaminationeffective}, and creating benchmarks to evaluate their performance \citep{chalkidis-etal-2022-lexglue,chalkidis-etal-2022-fairlex,niklaus-etal-2023-lextreme,guha2023legalbenchcollaborativelybuiltbenchmark,fei2023lawbenchbenchmarkinglegalknowledge,joshi2024ilturbenchmarkindianlegal,stern2024lawlanguagesbenchmarkingmultilingual,NEURIPS2022_d15abd14}. However, these efforts are often tailored to specific legal systems and jurisdictions, limiting their applicability to other legal contexts. Differences in the systems (e.g., civil law \textit{vs} common law) and the usual reliance on jurisdiction-specific laws mean that advances in one language or legal system are often not transferable to others.

European Portuguese has seen limited research, particularly in the legal field \citep{Rodrigues_2023,santos-etal-2024-advancing,lopes-etal-2024-gloria,10.1007/978-3-031-49011-8_12}, and no standardized benchmarks exist to evaluate LLMs specifically for Portuguese law. Some work on multilingual legal datasets exists \citep{chalkidis-etal-2021-multieurlex,aumiller-etal-2022-eur,niklaus-etal-2023-lextreme}, but they do not cover several important areas of the Portuguese law, and include only classification and summarization tasks. To address this, we introduce LegalBench.PT\footnote{Dataset publicly available at: \url{https://huggingface.co/datasets/BeatrizCanaverde/LegalBench.PT}}, the first benchmark that measures LLMs' legal knowledge and its practical application across key areas of Portuguese law.

We create LegalBench.PT by developing a taxonomy of the Portuguese law and collecting exams from a leading law school in Portugal. Since these exams rarely include multiple-choice questions and focus on long-form analysis, which is hard to evaluate automatically, we instruct GPT-4o \citep{Hello-GPT-4o} to convert the exam exercises into multiple-choice, true/false, and matching questions. Then, we filter the generated dataset to remove duplicates and undesirable instances, and shuffle the alternative options in multiple-choice and matching questions to minimize potential biases. The final dataset includes $4{,}723$ questions distributed across $31$ distinct legal areas. A subset of LegalBench.PT is reviewed by a lawyer, and, as expected with synthetic data, there is some noise: $12\%$ of answers are incorrect, and $15\%$ have suboptimal legal terminology or need rephrasing. We compare leading LLMs on LegalBench.PT, finding that GPT-4o and Claude-3.5-Sonnet \citep{claude-sonnet} are the strongest models, closely followed by Claude-3-Opus \citep{claude} and the open-source model Llama-3.1-405B \citep{dubey2024llama3herdmodels}. We analyze potential biases in GPT-4o’s responses, as this model generated the questions for the benchmark. By repeating the data creation process with Claude-3.5-Sonnet on a few key legal areas and evaluating both models on both datasets, we find no significant biases. Finally, we assess Portuguese lawyers on a sample of questions, and observe that their performances are usually closer to those of the lower-performing models Llama-3.1-8B and Mixtral-8x7B. This assessment highlights the presence of ambiguous questions and confirms the previously reported rate of incorrect gold answers.

\section{Related Work}

Legal evaluation datasets have traditionally focused on tasks that language models learn through fine tuning. These datasets, often derived from public online sources or expert annotations, include
tasks such as document review \citep{NEURIPS_DATASETS_AND_BENCHMARKS2021_6ea9ab1b,wang-etal-2023-maud}, judgment prediction \citep{chalkidis-etal-2019-neural,malik-etal-2021-ildc}, case
summarization \citep{NEURIPS2022_552ef803}, information extraction \citep{yao-etal-2022-leven}, among others \citep{chalkidis-etal-2022-lexglue,niklaus-etal-2023-lextreme,NEURIPS2022_d15abd14}. Although valuable, they do not fully capture the broader capabilities of LLMs in legal contexts.

Recent efforts have shifted towards developing benchmarks specifically for LLMs. MMLU \citep{hendrycks2021measuring}, an English multiple-choice test specifically designed for LLMs, includes a subset of legal questions
useful for preliminary assessments, but not always aligned with specific legal systems or jurisdictions. In contrast, professional certification exams offer more tailored
evaluations \citep{Zhong_Xiao_Tu_Zhang_Liu_Sun_2020,doi:10.1098/rsta.2023.0254,junior2024jurulegalbrazilianlarge}, but often fall short in comprehensively assessing LLMs’ practical use cases.

LegalBench \citep{guha2023legalbenchcollaborativelybuiltbenchmark} marks the first collaborative effort to benchmark legal reasoning for American law. It integrates existing and expert-crafted
datasets to assess practical legal reasoning skills, such as issue identification, rule applicability, and
text interpretation. This structured approach helps legal professionals understand the utility and limitations of models. Other benchmarks \citep{fei2023lawbenchbenchmarkinglegalknowledge,dai2024laiwchineselegallarge} similarly group tasks to separately evaluate legal knowledge, inference, and application.

\section{Data Collection}

The Portuguese law can be grouped into $5$ main areas:
1) \textbf{Public Law}: regulates relationships between private entities and the State; 
2) \textbf{Private Law}: governs relationships between private entities without State involvement; 
3) \textbf{Public-Private Law}: addresses issues spanning both Public and Private Law, depending on context; 
4) \textbf{Public International Law}: manages relations between states and international organizations; 
5) \textbf{EU and Community Law}: governs interactions between EU member states and EU institutions. 

\begin{figure}[ht]
\centering
  \includegraphics[width=0.9\columnwidth]{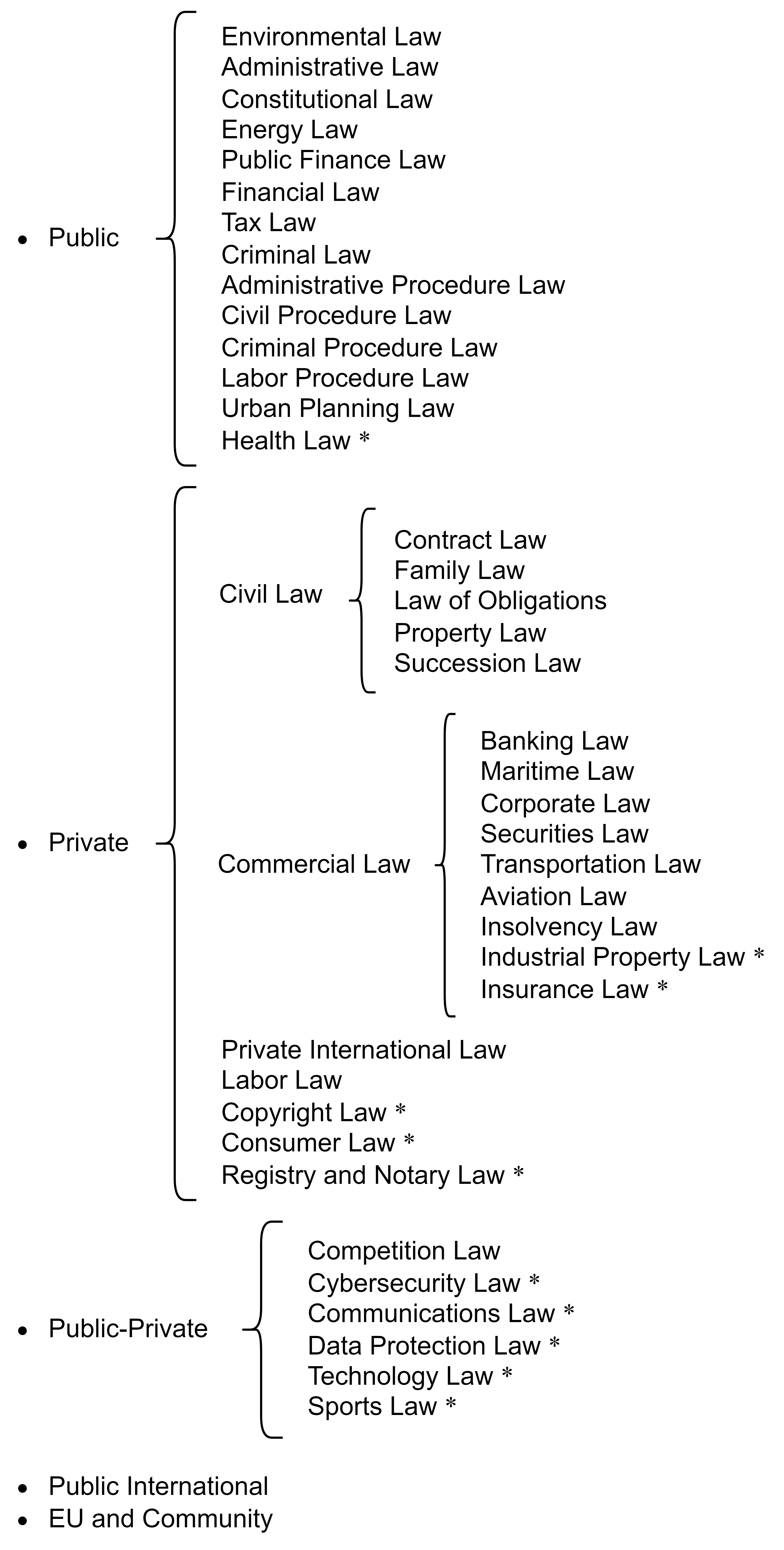}
  \caption{Full taxonomy of the Portuguese law adopted. Names marked with \** are not included in the benchmark due to lack of source data.}
  \label{fig:full-taxonomy}
\end{figure}

Figure~\ref{fig:full-taxonomy} shows the full taxonomy of the Portuguese law adopted. Public International and EU and Community stand alone without subareas.

\subsection{Data Processing}

We manually collected $341$ exams with solutions from the Faculty of Law at the University of Lisbon\footnote{\url{https://www.fd.ulisboa.pt/cursos/licenciatura/avaliacao/exames-escritos/}} covering the academic years 2021-2024. With a lawyer's help, we categorized them within our taxonomy. The exams, primarily composed of open-ended questions with rare multiple-choice items, required students to thoroughly analyze cases and take well-supported positions on legal issues. Downloaded as PDFs, we extracted the text using PyMuPDF\footnote{\url{https://pypi.org/project/PyMuPDF/}} and manually segmented the questions, separating independent questions and grouping related ones together. We also removed page numbers and headers, and placed in more convenient places footnotes appearing in the middle of the texts.

\subsection{Statistics}

We achieved broad coverage across several areas, with $13$ distinct fields in Public Law and $15$ in Private Law. In contrast, the Public-Private group is limited to a single field (Competition Law) due to the unavailability of suitable exams in other areas. Table~\ref{table:categories-questions-numbers} provides a breakdown of the number of distinct exam exercises across legal areas. Although specific subfields within Commercial Law were identified in our taxonomy, exams from courses named ``Commercial Law'' could not be mapped to more specific categories due to their broad content. As a result, Commercial Law is treated as a standalone field rather than an aggregate of its subfields. On the other hand, we do not report separate numbers for ``Civil Law'', as it is simply an aggregate of its subfields.

\begin{figure*}[ht]
\centering
\includegraphics[width=1.0\textwidth]{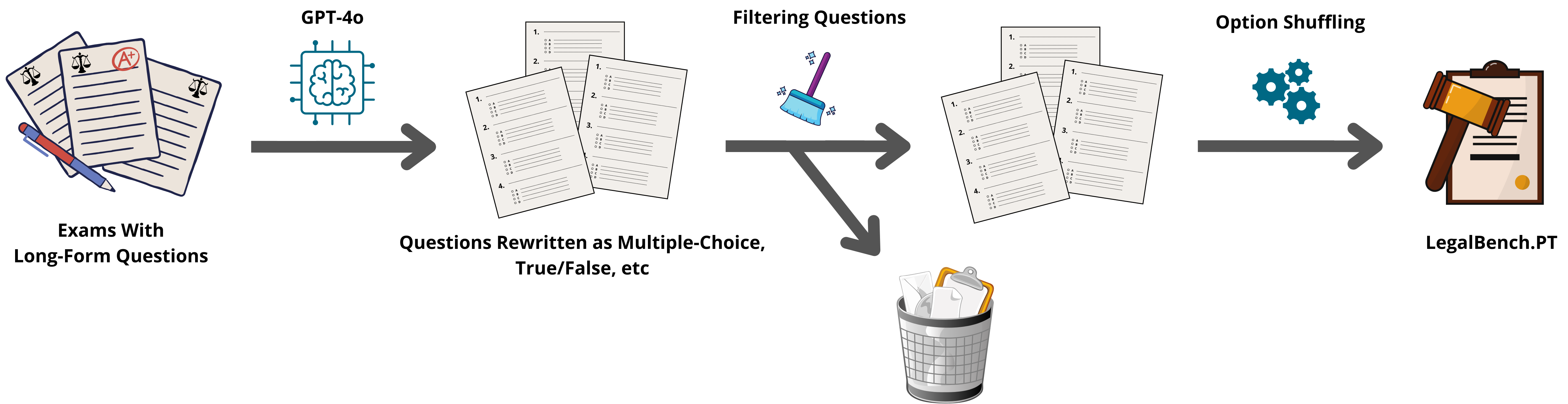}
\caption{LegalBench.PT construction pipeline.}
\label{fig:benchmark-pipeline}
\end{figure*}

\begin{table}[ht]
\scriptsize
\centering
\begin{tabular}{lcc}
\toprule
\textbf{Fields of Law} & \textbf{Exams} & \textbf{Benchmark} \\ 
\midrule
\textbf{Public} & 274 & 2115 \\ \midrule
Environmental Law & 11 & 52 \\
Administrative Law & 39 & 227\\
Constitutional Law & 58 & 206 \\
Energy Law & 12 & 21 \\
Public Finance Law & 25 & 208 \\
Financial Law & 1 & 16 \\
Tax Law & 19 & 232 \\
Criminal Law & 38 & 390 \\
Administrative Procedure Law & 19 & 116 \\
Civil Procedure Law & 20 & 272 \\
Criminal Procedure Law & 20 & 274 \\
Labor Procedure Law & 2 & 3 \\ 
Urban Planning Law & 10 & 98 \\
\midrule 
\textbf{Private} & 252 & 2057 \\ \midrule
Contract Law & 25 & 212 \\
Family Law & 34 & 220 \\
Law of Obligations & 17 & 218 \\
Property Law & 34 & 502 \\
Succession Law & 1 & 17 \\
Commercial Law & 23 & 240 \\
Banking Law & 2 & 30 \\
Maritime Law & 7 & 8 \\
Corporate Law & 12 & 154 \\
Securities Law & 9 & 45 \\ 
Transportation Law & 15 & 26 \\
Aviation Law & 19 & 31 \\
Insolvency Law & 4 & 37 \\
Private International Law & 18 & 110 \\
Labor Law & 32 & 207 \\
\midrule 
\textbf{Public-Private} (Competition Law) & 12 & 175 \\ \midrule
\textbf{Public International} & 72 & 212 \\
\midrule
\textbf{EU and Community} & 28 & 164 \\
\midrule 
\textbf{Total} & 626 & 4723 \\ \bottomrule
\end{tabular}
\caption{\label{table:categories-questions-numbers}
Distribution of distinct questions in the exams and the benchmark across fields of law.}
\end{table}

\section{Benchmark Creation}

In our initial experiments, we assessed GPT-4o's performance on long-form questions, following the methodology of \citet{NEURIPS2023_91f18a12}, where the model graded its own responses against the ``gold'' answers from the exams. However, this approach proved ineffective due to the detailed nature of the answers and the absence of clear error-penalization criteria. As a result, we converted the exam questions into more easily assessable formats: multiple-choice, true/false, and matching questions.

As Figure~\ref{fig:benchmark-pipeline} illustrates, our pipeline consists of three main phases: 1) providing GPT-4o with exam questions and answers to generate new question-answer pairs; 2) filtering the dataset to remove undesired questions using rules, and duplicates among questions with significant overlap; and 3) shuffling the alternative options from multiple-choice and matching questions to minimize biases and balance the distribution of the correct answers. Experiments were conducted with GPT-4o\footnote{Version: gpt-4o-2024-05-13.} \citep{Hello-GPT-4o} at a temperature of $0.001$, from May to August 2024.

\subsection{Question Generation}
\label{sec:question-generation}

To handle the different types of exam exercises, we implemented three approaches for generating new questions:
\begin{enumerate}
    \item \label{question-generation-third-approach} Providing the model with a group of short, independent questions, which typically ask for topic development or concept differentiation, along with their brief answers. We instruct the model to output new question-answer pairs.
    \item \label{question-generation-second-approach} Feeding the model one question-answer pair at a time, useful for long, complex questions and answers. These questions are usually long because they include a problem statement, often a case analysis. We instruct the model to identify the problem statement and generate new question-answer pairs.
    \item \label{question-generation-first-approach} Presenting the model with a set of exam questions and answers related to a common problem statement. These exam questions sometimes present new assumptions, as a continuation of the problem statement, that may contain critical information. We request the model to identify the problem statement and assumptions, and generate new question-answer pairs.
\end{enumerate}

Within each field of law, questions were divided into three groups, corresponding to the three approaches. Each exam question-answer pair was shown to GPT-4o twice: in the second round, the model was also shown the respective question-answer pairs generated in the first round. This was particularly necessary for approaches \ref{question-generation-second-approach} and \ref{question-generation-first-approach} because some answers were lengthy, and GPT-4o's output token limit of $4{,}096$ often caused incomplete coverage of all topics. We designed six prompts in total, two for each approach.

We specified the types of questions to be generated: 1) \textbf{multiple-choice}, with only one correct option; 2) \textbf{cloze tasks}, fill-in-the-blank exercises formulated as multiple-choice; 3) \textbf{case analysis}, multiple-choice questions where the model must create a small case for analysis; 4) \textbf{true/false}, requiring classification of statements as either ``True'' or ``False''; 5) \textbf{multiple selection}, multiple-choice questions where more than one option can be correct; 6) \textbf{matching questions}, requiring respondents to pair items from two columns. Case analysis questions were only requested for approach \ref{question-generation-third-approach}, as the exam questions were very theoretical. Matching questions were only requested for approaches \ref{question-generation-second-approach} and \ref{question-generation-first-approach}, since preliminary experiments showed that those generated with approach \ref{question-generation-third-approach} used to test the knowledge of specific laws and articles, an issue that we will discuss in Section \ref{sec:filtering}. The remaining types were requested for all approaches.

After generation, to get the final version of the new questions, we joined them together with the respective statements and assumptions. For approach \ref{question-generation-second-approach}, we join the statement at the beginning of each question. For approach \ref{question-generation-first-approach}, we join, at the beginning of each question, the statement and all assumptions preceding the question. This method ensures that important information from previous assumptions is not lost. However, in the filtering and shuffling steps in Sections~\ref{sec:filtering} and~\ref{sec:option-shuffling}, we focus on the questions as extracted from the outputs, without considering the problem statements and assumptions. See Appendix~\ref{sec:question-generation-details} for examples.

\subsection{Filtering}
\label{sec:filtering}

After generation, we reviewed random outputs from the different fields of law and identified two main issues: questions referencing specific articles and laws, and repeated questions.

\paragraph{\textbf{Articles and Laws}} Many questions referenced specific articles and laws by number (e.g., ``Artigo 103.º, n.º 2 da Constituição''\footnote{Article 103.º, n.º 2 of the Constitution.}) to test the model's knowledge of legal texts. We removed these questions for two reasons. First, laws and articles are frequently updated, and their relevance depends on the timing of events. Since some questions and statements did not specify dates, they could be misleading. Second, lawyers are not expected to memorize legal provisions. We removed all questions that matched both of the following patterns in lowercase: \verb/(?<!^)(?<!\n)\d+/ and \verb/n\.º|nº|[^a-z](artigos?|art|reg)[^a-z]/.
 
\paragraph{\textbf{Repeated Questions}} We found a significant number of repeated or similar questions within batches generated from the same exam exercise(s) fed into GPT-4o (first $+$ second rounds). Given each batch, we compared multiple-choice, cloze tasks, case analysis, multiple selection, true/false, and matching questions. We first compared questions within each of these types, and then compared questions of different types. The first four are multiple-choice variants: for these, we compared both the questions alone and with answer options, which allowed us to detect similarities at both the question and content levels. True/false questions are simply sentences to be classified as true or false, so we did not process them. In some cases, matching questions were processed into a more convenient format: we removed the first line (usually ``Match the items...'') and the letters/numbers identifying options (see Appendix~\ref{sec:filtering-repeated-questions} for an example).

To remove repeated questions, we used a combination of lexical and semantic methods for filtering. For each pair of questions, we selected the most appropriate ROUGE-L variant (sentence- or summary-level), and, after processing the questions, we used the Python implementation of ROUGE-L\footnote{\url{https://pypi.org/project/rouge-score/}} to compute precision and recall scores. If $min(precision, recall)$ met or exceeded the threshold, we eliminated one of the questions. Additionally, we used a multilingual Transformer model trained to detect semantically similar sentences\footnote{\url{https://huggingface.co/sentence-transformers/paraphrase-multilingual-mpnet-base-v2}} to compute a sentence embedding for each question. For each pair of embeddings, we measured their cosine similarity and if it met or exceeded the threshold, we removed one of the questions. A summary of the methods used, question processing, and thresholds used for filtering within and between types is available in Appendix~\ref{sec:filtering-repeated-questions}. We manually set the best threshold for each scenario.

Before comparing questions within a group, we randomized their order to avoid altering the distribution across types. Once a question met the elimination criteria, it was removed from further comparisons.

\subsection{Option Shuffling}
\label{sec:option-shuffling}

After filtering, we analyzed the frequency of the correct answer options within multiple-choice, cloze tasks, case analysis, and multiple selection categories. For reference, $98.2\%$ of these questions had $4$ answer options, with the remainder having between $2$ and $7$ options. We found a highly imbalanced distribution, with GPT-4o exhibiting a pronounced tendency to generate questions where the second option was the correct one.

To minimize potential biases at evaluation time and balance the answer choices, we randomized the order of the options in all these question types. Upon inspection, we encountered $18$ different questions from multiple-choice variants with answer options such as ``Both options a) and b) are possible.'', ``None of the above.'', and ``All of the above.''\footnote{``Ambas as opções a) e b) são possíveis.'', ``Nenhuma das anteriores.'', ``todas as anteriores''}. To handle these, we implemented two approaches: 
\begin{itemize}
    \item For each answer option, we used the pattern \verb/\s+[a-z]\)\s*/ to check for references to other options. If a match was found, we kept the question as it was originally created.
    \item For each answer option, we checked if it contained the word ``anteriores'' along with either ``nenhum'' or ``todas''\footnote{Contextual English translation: we checked if it contained the word ``above'' along with either ``none'' or ``all''.}. Any option matching these words was always the last option of the question. In these cases, we shuffled only the preceding answer options.
\end{itemize}
For the remaining questions, we shuffled all the options. We also shuffled the items for matching questions but did not account for exceptions, as we did not find any.

\subsection{Dataset Statistics}

We obtained a total of $4{,}723$ new questions (down from $10{,}951$ before filtering), as shown in Table~\ref{table:categories-questions-numbers}. The differences between the distribution of the exam exercises and benchmark questions is large. This is mainly caused by the number of exam questions run with each generation approach outlined in Section~\ref{sec:question-generation}. While an exam question with a long answer is usually converted into multiple new questions, a short exam answer is usually converted into only one or two questions.

Out of $31$ fields, $14$ contain over $200$ questions each, and $6$ have between $98$ and $200$ questions, covering nearly all fundamental fields of Portuguese law and representing $94\%$ of the benchmark. The remaining $11$ areas have fewer than $60$ questions. Of these, $7$ contain between $20$ and $60$ questions, offering a moderate assessment. The remaining $4$ fields have fewer than $20$ questions, making them insufficient for a comprehensive evaluation. Nonetheless, we include these questions as they provide some value and to potentially aid future research. See Appendix~\ref{sec:more-statistics} for further details.

\subsection{Question Analysis}
\label{sec:question-analysis}

One of the authors, a practicing lawyer, reviewed $2$ small samples of questions and confirmed their relevance and usefulness in assessing a model’s legal knowledge. They also identified a few lower-quality questions, which we discuss below.

\paragraph{\textbf{Corrections}} We randomly selected two groups of questions from different fields\footnote{EU and Community, and Civil Procedure Law.} for analysis. Out of $33$ questions, $4$ contained incorrect answers, and $5$ required minor improvements, such as correcting legal terminology or rephrasing for clarity.

\paragraph{\textbf{``Easy'' Questions}} In a sample of $64$ statement-based questions across three fields\footnote{EU and Community, Civil Procedure, Competition Law.} ($19$ questions from the previous sample, $45$ randomly chosen), $12$ primarily tested interpretation skills rather than legally relevant content. We considered these questions too straightforward and tried to filter them out with GPT-4o, with no success.

\subsection{Comparison to Other Works}

Unlike other legal benchmarks \citep{guha2023legalbenchcollaborativelybuiltbenchmark,fei2023lawbenchbenchmarkinglegalknowledge,dai2024laiwchineselegallarge}, LegalBench.PT does not offer a task-specific measure of LLMs performance in usual tasks, such as judgement prediction, summarization, or information extraction.

LegalBench \citep{guha2023legalbenchcollaborativelybuiltbenchmark} is organized according to five different types of legal reasoning: issue-spotting, rule-recall, rule-conclusion, interpretation, and rhetorical-understanding. LegalBench.PT includes questions that fit in these categories, but is not organized according to them. The authors of LegalBench argue that their legal framework provides lawyers and LLM developers with a common vocabulary, which is fundamental for enabling a cross-disciplinary understanding of LLM capabilities in the legal domain. Such categorization can be performed in future work.

LegalBench.PT offers a similar solution to works that use professional certification exams \citep{junior2024jurulegalbrazilianlarge,Zhong_Xiao_Tu_Zhang_Liu_Sun_2020,doi:10.1098/rsta.2023.0254}, which typically focus on multiple-choice questions. It provides: 1) the same straightforward assessment through multiple-choice and other similar formats, which relieves models from the need to generate long, detailed explanations that complicate an accurate assessment; and 2) a tailored evaluation of legal expertise specific to the Portuguese law.

\section{Evaluation}

Our evaluations are conducted in a zero-shot setting. The prompts always specify the legal area, provide detailed instructions on the type of question (e.g., whether there is one or multiple correct answers), and outline the expected response format. An example template can be found in Appendix~\ref{sec:prompts-evaluation}.

\begin{table*}[ht]
\scriptsize
\centering
\begin{tabular}{lcccccccc}
\toprule
\multirow{2}{*}{\textbf{\begin{tabular}[c]{@{}c@{}}Fields of Law\\ \end{tabular}}} & \multirow{2}{*}
{\textbf{\begin{tabular}[c]{@{}c@{}}GPT-\\ 4o\end{tabular}}} & \multirow{2}{*}
{\textbf{\begin{tabular}[c]{@{}c@{}}GPT-\\ 4o-mini\end{tabular}}} & \multirow{2}{*}
{\textbf{\begin{tabular}[c]{@{}c@{}}Claude-\\ 3-Opus\end{tabular}}} & \multirow{2}{*}
{\textbf{\begin{tabular}[c]{@{}c@{}}Claude-\\ 3.5-Sonnet\end{tabular}}} & \multirow{2}{*}
{\textbf{\begin{tabular}[c]{@{}c@{}}Llama-\\ 3.1-8B\end{tabular}}} & \multirow{2}{*}
{\textbf{\begin{tabular}[c]{@{}c@{}}Llama-\\ 3.1-70B\end{tabular}}} & \multirow{2}{*}
{\textbf{\begin{tabular}[c]{@{}c@{}}Llama-\\ 3.1-405B\end{tabular}}} & \multirow{2}{*}
{\textbf{\begin{tabular}[c]{@{}c@{}}Mixtral-\\ 8x7B\end{tabular}}} \\
 &  &  &  &  &  &  \\ \midrule
\textbf{Public} & 84.4 & 78.3 & 82.5 & 84.2 & 65.8 & 79.4 & 81.8 & 66.5 \\ 
\midrule
Environmental Law & 94.8 & 85.0 & 88.0 & 89.9 & 71.3 & 87.7 & 89.8 & 76.4 \\
Administrative Law & 86.9 & 80.4 & 82.7 & 88.7 & 64.4 & 80.2 & 81.3 & 66.8 \\
Constitutional Law & 88.0 & 81.9 & 86.8 & 89.2 & 67.5 & 85.4 & 87.1 & 70.1 \\
Energy Law & 94.3 & 86.0 & 79.2 & 81.4 & 71.5 & 83.8 & 80.6 & 89.2 \\
Public Finance Law & 87.6 & 83.5 & 87.2 & 88.8 & 73.3 & 86.8 & 89.3 & 72.9 \\
Financial Law & 95.6 & 87.2 & 96.5 & 99.1 & 75.3 & 94.2 & 95.6 & 75.6 \\
Tax Law & 89.1 & 77.2 & 87.8 & 90.1 & 58.0 & 76.0 & 80.0 & 60.0 \\
Criminal Law & 79.0 & 71.7 & 77.9 & 79.4 & 62.7 & 73.0 & 77.6 & 61.8 \\
Administrative Procedure Law & 80.5 & 81.1 & 78.8 & 80.2 & 68.5 & 78.9 & 81.1 & 68.2 \\
Civil Procedure Law & 79.8 & 75.2 & 78.8 & 78.0 & 65.2 & 77.4 & 78.5 & 66.6 \\
Criminal Procedure Law & 80.9 & 73.9 & 76.7 & 77.7 & 62.9 & 74.9 & 76.6 & 59.3 \\
Labor Procedure Law & 66.7 & 66.7 & 66.7 & 66.7 & 33.3 & 66.7 & 33.3 & 0.0 \\ 
Urban Planning Law & 93.9 & 94.2 & 94.6 & 95.3 & 82.4 & 94.8 & 95.6 & 87.1 \\
\midrule
\textbf{Private} & 84.8 & 80.2 & 84.3 & 84.1 & 69.1 & 79.9 & 83.9 & 70.6 \\
\midrule
Contract Law & 81.2 & 77.1 & 81.7 & 82.0 & 68.2 & 77.6 & 81.3 & 64.9 \\
Family Law & 81.3 & 72.9 & 82.6 & 81.6 & 64.3 & 77.4 & 77.7 & 65.0 \\
Law of Obligations & 84.0 & 76.3 & 79.8 & 80.7 & 61.1 & 78.7 & 80.6 & 65.8 \\
Property Law & 85.1 & 79.7 & 84.0 & 83.8 & 69.6 & 79.4 & 84.2 & 70.4 \\
Succession Law & 60.4 & 66.9 & 72.9 & 75.9 & 56.9 & 58.3 & 62.7 & 45.9 \\
Commercial Law & 88.6 & 82.5 & 89.9 & 86.3 & 69.8 & 78.6 & 85.3 & 74.7 \\
Banking Law & 98.3 & 97.7 & 98.3 & 99.2 & 95.3 & 96.6 & 96.9 & 90.7 \\
Maritime Law & 87.5 & 60.0 & 87.5 & 87.5 & 68.8 & 56.2 & 75.0 & 87.5 \\
Corporate Law & 79.0 & 79.2 & 79.3 & 79.1 & 67.9 & 77.1 & 80.8 & 69.2 \\
Securities Law & 92.9 & 89.2 & 84.1 & 85.7 & 77.1 & 89.7 & 91.2 & 78.9 \\ 
Transportation Law & 67.2 & 88.9 & 70.2 & 84.7 & 48.2 & 74.4 & 77.0 & 63.6 \\
Aviation Law & 91.5 & 85.9 & 89.5 & 87.0 & 67.9 & 90.9 & 85.1 & 70.6 \\
Insolvency Law & 71.7 & 85.9 & 75.4 & 80.9 & 73.7 & 68.5 & 84.4 & 66.1 \\
Private International Law & 86.6 & 82.7 & 90.0 & 89.1 & 71.2 & 83.7 & 92.1 & 71.9 \\
Labor Law & 93.1 & 87.3 & 89.9 & 89.9 & 78.1 & 87.9 & 91.4 & 81.5 \\
\midrule
\textbf{Public-Private} (Competition Law) & 93.6 & 93.7 & 94.4 & 95.0 & 87.9 & 94.5 & 95.5 & 81.2 \\
\midrule
\textbf{Public International} & 89.6 & 84.5 & 90.5 & 90.4 & 75.4 & 86.5 & 90.4 & 77.2 \\
\midrule
\textbf{EU and Community} & 91.3 & 83.7 & 90.4 & 92.8 & 69.4 & 85.3 & 89.0 & 77.0 \\
\midrule
\textbf{Overall} & 85.4 & 80.1 & 84.3 & 85.1 & 68.6 & 80.7 & 83.8 & 69.7 \\
\bottomrule
\end{tabular}
\caption{\label{table:models-results-categories}
Model performance ($\%$) across the different fields of law.}
\end{table*}

From the model outputs, we extract letter options, true/false classifications, and matching pairs according to the instructed format. We evaluate multiple-choice, cloze tasks, case analysis, and true/false questions separately using balanced accuracy. For multiple selection and matching questions, we use the $F_1$ score. We aggregate the results from the different quesion types and fields of law using a weighted average.

\subsection{Model Performance}

We evaluated on LegalBench.PT: GPT-4o\footnote{Version: gpt-4o-2024-05-13.} \citep{Hello-GPT-4o}, GPT-4o-mini\footnote{Version: gpt-4o-mini-2024-07-18.} \citep{GPT-4o-mini-advancing-cost-efficient-intelligence}, Claude-3-Opus\footnote{Version: claude-3-opus-20240229.} \citep{claude}, Claude-3.5-Sonnet\footnote{Version: claude-3-5-sonnet-20240620.} \citep{claude-sonnet}, Llama-3.1-8B, Llama-3.1-70B, Llama-3.1-405B \citep{dubey2024llama3herdmodels}, and Mixtral-8x7B \citep{jiang2024mixtralexperts}. We used the instruct versions of the open-source models. We set the temperature to $0.01$. Evaluations were conducted in September $2024$.

Table~\ref{table:models-results-categories} shows the models performance, in percentage, across the various legal areas. Overall, GPT-4o is the best-performing model, closely followed by Claude-3.5-Sonnet. Notably, the open-source model Llama-3.1-405B trails behind GPT-4o with a difference of only $1.6\%$, and the smaller version Llama-3.1-70B achieves results similar to GPT-4o-mini across most areas.
Apart from Llama-3.1-8B and Mixtral-8x7B, which show noticeably weaker results, likely due to their smaller size, the remaining models have
similar performances. This suggests either similar capabilities or that LegalBench.PT is not challenging enough to differentiate these models.

Certain legal fields, such as Labor Procedure and Succession Law, consistently show lower scores, likely due to their smaller question sets. Similarly, Maritime, Transportation, Environmental, Energy, Financial, Banking, Aviation, and Insolvency Law have small question sets, making their results less informative, as they do not represent a comprehensive evaluation. Conversely, certain areas, namely Urban Planning, Banking, Labor, and Competition Law consistently show high performance across all models, likely due to less challenging questions.

The higher performance in the Public-Private area may not be fully comparable to the broader Public and Private Law fields due to the smaller and less diverse question set, which encompasses only Competition Law. Compared to the Public and Private areas, all models perform better in Public International and EU and Community Law, likely because those areas involve smaller and less diverse question sets, and cover topics more common across multiple jurisdictions and languages, which are likely better represented in the training data. Appendix~\ref{sec:appendix-models-performance} presents the models performance across the different question types.

\subsection{Investigating GPT-4o Bias}

Table~\ref{table:models-results-categories} shows that GPT-4o is one of the top-performing models, with overall results similar to Claude-3.5-Sonnet. Since GPT-4o was used to generate the dataset, it can be argued that this comparison is unfair. To test this hypothesis, we repeated the data creation process (using the same prompts, filtering, and shuffling) with Claude-3.5-Sonnet. To facilitate the analysis, we focused on $6$ key areas\footnote{Administrative, Civil Procedure, Family, Commercial, Public International, EU and Community Law}, generating $1{,}389$ questions, slightly more than the $1{,}335$ questions in the original dataset for these areas, and similarly distributed.

Table~\ref{table:results-gpt-vs-claude} compares GPT-4o and Claude-3.5-Sonnet on these datasets. Claude-3.5-Sonnet outperforms GPT-4o in all areas except EU and Community Law, with performance differences ranging from $0.6\%$ to $3.5\%$. In the GPT-generated data, we observe a similar behaviour, with Claude-3.5-Sonnet surpassing GPT-4o in $4$ areas, whose differences range from $0.3\%$ to $1.8\%$. However, in the other $2$ areas GPT-4o surpasses Claude-3.5-Sonnet by $1.8\%$ and $2.3\%$. Overall, models show identical performance on the GPT-4o-generated questions and a small difference on the Claude-generated ones. It is possible that both models may have a slight advantage when handling their own generated data, but the difference seems negligible.

\begin{table}[ht]
\scriptsize
\centering
\begin{tabular}{lcccc}  \cmidrule[\heavyrulewidth]{2-5}
& \multicolumn{2}{c}{\textbf{GPT Generated}} & \multicolumn{2}{c}{\textbf{Claude Generated}}
\\ \cmidrule[\heavyrulewidth]{1-1} \cmidrule(lr){2-3} \cmidrule(lr){4-5}
\textbf{Fields of Law} & \textbf{GPT} & \textbf{Claude} & \textbf{GPT} & \textbf{Claude} \\ \midrule
Administrative Law & 86.9 & 88.7 & 82.3 & 85.8 \\
Civil Procedure Law & 79.8 & 78.0 & 80.3 & 81.4 \\
Family Law & 81.3 & 81.6 & 84.9 & 87.0 \\
Commercial Law & 88.6 & 86.3 & 78.8 & 82.1 \\
Public International Law & 89.6 & 90.4 & 91.6 & 92.2 \\
EU and Community Law & 91.3 & 92.8 & 92.8 & 91.6 \\ \midrule
\textbf{Overall} & 85.8 & 85.7 & 84.5 & 86.2 \\ \bottomrule
\end{tabular}
\caption{GPT-4o and Claude-3.5-Sonnet performance ($\%$) on the questions generated by each of these models across different fields of law.}
\label{table:results-gpt-vs-claude}
\end{table}

We conducted a Wilcoxon signed-rank test to assess whether the differences between the models' performances on the datasets corresponding to different fields of law follow a symmetric distribution around zero. For the GPT-generated data, we obtained a $p$-value of $100\%$, and for the Claude-generated data, a $p$-value of $15.6\%$. In both cases, there is no evidence for considering the performance differences between GPT-4o and Claude-3.5.-Sonnet statistically significant.

\subsection{Human Evaluation}

We conducted a human evaluation with Portuguese lawyers on LegalBench.PT to establish a baseline for model comparison and validate the benchmark. Given the high cost of a lawyer's time and the large number of questions in our benchmark, we randomly selected $1{,}000$ questions from $10$ fundamental areas of law ($100$ questions from each area)\footnote{Administrative, Constitutional, Criminal, Civil Procedure, Contract, Family, Commercial, Labor, Public International, and EU and Community Law}. To minimize the effort and time required from each lawyer, we randomly divided these questions into $20$ groups, each containing $50$ questions, and assigned each lawyer at least one group.

We deployed the survey using \texttt{streamlit}\footnote{\url{https://streamlit.io/}}. We slightly simplified the prompts used to evaluate LLMs, and presented them to the lawyers one question at a time. To simplify the answering process, multiple-choice, cloze tasks, case analysis, and true/false questions used checkboxes, allowing participants to easily select the correct option. We provided the participants with guidelines allowing them to consult legal texts, books, the internet, or other resources, but they were prohibited from seeking opinions from others or using language models. 

In total, $22$ lawyers participated, providing $1{,}183$ answers across $17$ different groups. Only two lawyers answered two groups each, while the remaining participants answered just one group each. For $350$ questions across $7$ groups, we collected paired answers, meaning each of these questions has answers from two lawyers. The remaining questions have a single response each. All groups have $50$ answered questions, except for one group that is reduced to $33$ questions because the lawyer who started answering it did not finish.

\subsubsection{Performance}

\noindent We analyzed each group individually, comparing the performance of the lawyers with that of the top-performing models (GPT-4o and Claude-3.5-Sonnet) and lower-performing models (Llama-3.1-8B and Mixtral-8x7B). The results, presented in Table~\ref{table:performance-people-models}, show that the lawyers' performance is usually closer to that of Llama-3.1-8B and Mixtral-8x7B, or worse, rather than matching or surpassing the results of GPT-4o and Claude-3.5-Sonnet. Only four people outperform GPT-4o or Claude-3.5-Sonnet (groups $6$, $7$, $11$). Additionally, in some groups the lawyers’ performance is significantly worse than any of the models (groups $4$, $8$, $16$).

\begin{table}[ht]
\tiny
\centering
\begin{tabular}{c|cccccc}
\toprule
\multirow{2}{*}{\textbf{Groups}} & \multirow{2}{*}
{\textbf{\begin{tabular}[c]{@{}c@{}}Person\\ 1\end{tabular}}} & \multirow{2}{*}
{\textbf{\begin{tabular}[c]{@{}c@{}}Person\\ 2\end{tabular}}} & \multirow{2}{*}
{\textbf{\begin{tabular}[c]{@{}c@{}}GPT\\ -4o\end{tabular}}} & \multirow{2}{*}{\textbf{\begin{tabular}[c]{@{}c@{}}Claude-3.5\\ -Sonnet\end{tabular}}} & \multirow{2}{*}{\textbf{\begin{tabular}[c]{@{}c@{}}Llama-\\ 3.1-8B\end{tabular}}} & \multirow{2}{*}{\textbf{\begin{tabular}[c]{@{}c@{}}Mixtral-\\ 8x7B\end{tabular}}} \\
 &  &  &  &  &  &  \\ \midrule
\textbf{1} & 79.0 & 83.3 & 94.4 & 88.5 & 75.3 & 77.9 \\
\textbf{2} & 73.7 & 66.9 & 82.5 & 81.0 & 68.3 & 75.9 \\
\textbf{3} & 61.7 & 80.3 & 87.8 & 87.4 & 60.6 & 66.6 \\
\textbf{4} & 59.1 & 67.1 & 91.6 & 95.3 & 80.6 & 81.5 \\
\textbf{5} & 66.3 & 84.2 & 92.8 & 88.2 & 63.4 & 71.5 \\
\textbf{6} & 65.8 & 91.4 & 88.1 & 87.0 & 61.7 & 72.2 \\
\textbf{7} & 77.1 & 75.6 & 74.8 & 75.9 & 67.6 & 74.7 \\
\textbf{8} & 56.1 & - & 71.7 & 78.2 & 72.7 & 69.1 \\
\textbf{9} & 67.2 & - & 80.6 & 86.4 & 69.3 & 73.1 \\
\textbf{10} & 57.8 & - & 95.2 & 93.5 & 68.2 & 59.5 \\
\textbf{11} & 82.9 & - & 82.0 & 85.8 & 71.9 & 72.1 \\
\textbf{12} & 83.0 & - & 88.4 & 88.2 & 83.1 & 77.3 \\
\textbf{13} & 64.1 & - & 75.1 & 83.5 & 62.4 & 54.8 \\
\textbf{14} & 61.3 & - & 73.4 & 74.3 & 65.0 & 56.3 \\
\textbf{15} & 77.0 & - & 79.7 & 82.4 & 56.1 & 59.7 \\
\textbf{16} & 60.0 & - & 90.9 & 89.0 & 73.5 & 78.4 \\
\textbf{17} & 78.4 & - & 88.8 & 90.8 & 74.7 & 72.0 \\ \bottomrule
\end{tabular}
\caption{\label{table:performance-people-models}
Human and model performance ($\%$) on disjoint groups of questions from LegalBench.PT. Person 1 and Person 2 differ across groups. All groups have $50$ questions randomly selected from different legal areas, except group $16$ which has only $33$ questions.}
\end{table}

Examining groups $1$ to $7$, we see some performance disparities between participants within the same group. While groups $1$, $2$, $4$, and $7$ show differences of less than $10\%$, groups $3$, $5$, and $6$ exhibit larger gaps of $18.6\%$, $17.9\%$, and $25.6\%$, respectively. Group $4$ stands out for both participants' low performance, contrasting sharply with the high scores of all models in this group. 

We reached out to person $1$ from group $5$ and person $1$ from group $6$. Each of these participants answered $50$ questions, making a total of $100$ questions. Out of these $100$ questions, $32$ were answered incorrectly according to our gold standards. We showed these participants the correct solutions to the questions they had answered incorrectly and received feedback on a total of $13$ questions, which can be summarized as follows:
\begin{itemize}
    \item $6$ questions were considered ambiguous. This means that their gold standards are correct, but so are the lawyers' answers. In these cases, different answers can be valid depending on how the questions are interpreted and the legal arguments used.
    \item For $7$ questions, the gold answers were indeed correct but participants answered incorrectly due to distractions. They mentioned that answering several questions consecutively led to overlooked details. We had previously received feedback about some questions being lengthy and complex. This likely contributed to distractions, especially for a human answering $50$ questions in a row.
\end{itemize}

Assuming that correctly answered questions are not ambiguous, we can estimate that $15$ questions ($14.8\%$) of the total $100$ were ambiguous. Due to lack of time, we could not reach out to all participants and study the causes of the low results in depth. In future work, it may be beneficial to include control questions in this type of assessment to help identify recurring distractions and provide a clearer understanding of their impact on results. Additionally, participants could be shown the correct solutions as they respond to the questions and asked to comment on any discrepancies between their answers and the gold standard. However, this would require human effort to analyze the responses.

\subsubsection{Agreement}

For groups with two participants (groups $1$ to $7$), we computed the agreement between the participants' answers. For each group, we iterated over multiple-choice, cloze tasks, case analysis, and true/false questions, and computed the percentage of questions where the participants agree. For multiple selection and matching questions, we used Jaccard similarity to compute the agreement for each pair of questions, and averaged across all pairs. To obtain the overall agreement for each group, we aggregated the scores across the different question types using a weighted average. We also computed the percentage of the agreed-upon answers that were correct according to the gold standards.

The results are shown in Table~\ref{table:agreement-groups-performance}. The agreement values are relatively low, averaging $62.4\%$. They are not surprising given the performance results shown in Table~\ref{table:performance-people-models}. Group $4$ has significantly lower agreement, which aligns with the previously mentioned lower performances. Questions on which participants disagree likely represent, at least in part, ambiguous questions.

\begin{table}[ht]
\footnotesize
\centering
\begin{tabular}{ccc}
\toprule
\multirow{2}{*}
{\textbf{\begin{tabular}[c]{@{}c@{}}Groups\\ \end{tabular}}} & \multirow{2}{*}
{\textbf{\begin{tabular}[c]{@{}c@{}}Agreement\\ (\%)\end{tabular}}} & \multirow{2}{*}
{\textbf{\begin{tabular}[c]{@{}c@{}}Accuracy\\ (\%)\end{tabular}}} \\ 
 &  &  \\ \midrule
1 & 69.7 & 94.3 \\ 
2 & 64.0 & 79.7 \\ 
3 & 57.8 & 89.7 \\ 
4 & 45.8 & 88.1 \\ 
5 & 63.6 & 91.7 \\ 
6 & 63.8 & 88.2 \\ 
7 & 71.8 & 89.7 \\ \midrule
\textbf{Overall} & 62.4 & 88.9 \\
\bottomrule
\end{tabular}
\caption{\label{table:agreement-groups-performance}
Agreement rates between participants' answers and accuracy of the agreed-upon answers on disjoint groups of questions from LegalBench.PT. Participants differ across groups.}
\end{table}

The high accuracy scores on the agreed-upon answers suggests that the respective gold standards are correct. Groups $3$ to $7$ show an error rate between $8\%$ and $12\%$, and group $1$ shows an even lower rate of just $5.7\%$. However, group $2$ presents a discrepant error of $20.3\%$. The errors may indicate incorrect gold answers. In Section~\ref{sec:question-analysis}, the lawyer who reviewed a sample of the generated questions identified $12.1\%$ questions with incorrect gold answers. This value closely matches the average error rate of $11.1\%$ observed in the current analysis. Since this lower value results from an assessment on a larger and more diverse set of questions, it might be a more accurate approximation of the true rate of incorrect gold answers in LegalBench.PT. On the other hand, we also recall the estimated $14.8\%$ of ambiguous questions. Although an ambiguous question does not necessarily indicate an incorrect gold answer, it does not allow a clear and fair assessment, as we saw in the previous section. This latter value, being close to $11.1\%$ and $12.1\%$, supports our estimate of the percentage of questions that we consider ``problematic''.

We also organized the questions by legal fields rather than by groups. Table~\ref{table:agreement-law-fields-performance} shows the number of questions per field, the agreement between participants in each field, and the accuracy of the agreed-upon answers. Each field includes questions from all seven groups, except Public International Law and EU and Community Law, which only contain questions from six groups. There is some variation in the agreement values across the fields. Half of these have an agreement rate below $60\%$, a pattern we only observed in two groups in Table~\ref{table:agreement-groups-performance}. On the other hand, the accuracy scores are more consistent. Administrative Law presents a slightly lower accuracy of $80.2\%$, while the remaining fields have accuracies between $86\%$ and $96\%$. If the error rate indeed represents incorrect gold standards, this suggests that faulty answers exhibit a balanced distribution across the fields, with only Administrative Law standing out due to a slightly higher proportion of incorrect answers.

\begin{table}[ht]
\scriptsize
\centering
\begin{tabular}{lccc}
\toprule
\multirow{2}{*}{\textbf{Fields of Law}} & \multirow{2}{*}
{\textbf{\begin{tabular}[c]{@{}c@{}}Total\\ Questions\end{tabular}}} & \multirow{2}{*}
{\textbf{\begin{tabular}[c]{@{}c@{}}Agreement\\ (\%)\end{tabular}}} & \multirow{2}{*}
{\textbf{\begin{tabular}[c]{@{}c@{}}Accuracy\\ (\%)\end{tabular}}} \\ 
 &  &  &  \\ \midrule 
Administrative Law & 40 & 68.3 & 80.2 \\ 
Constitutional Law & 39 & 61.5 & 95.4 \\ 
Criminal Law & 40 & 56.9 & 86.0 \\ 
Civil Procedure Law & 34 & 58.8 & 87.5 \\ 
Contract Law & 34 & 54.9 & 91.3 \\ 
Family Law & 33 & 57.4 & 87.0 \\ 
Commercial Law & 35 & 70.2 & 89.9 \\ 
Labor Law & 38 & 71.1 & 93.1 \\ 
Public International Law & 23 & 56.1 & 91.7 \\ 
EU and Community Law & 34 & 65.2 & 88.5 \\ \midrule 
\textbf{Overall} & 350 & 62.4 & 88.9 \\ 
\bottomrule
\end{tabular}
\caption{\label{table:agreement-law-fields-performance}
Number of questions analyzed in each field, agreement rates between participants' answers to those questions, and accuracy of the agreed-upon answers. Each field includes responses from $6$ to $7$ different participant pairs.}
\end{table}

\section{Conclusion}

We introduced LegalBench.PT, the first legal benchmark designed to evaluate LLMs' knowledge and application of Portuguese law. Organized according to a taxonomy of the Portuguese law, it allows for the measurement and comparison of LLMs' proficiency across various legal fields.

We demonstrated that synthetic data generation, grounded on law school exams and with adequate post-processing and filtering, can produce a good benchmark for model evaluation. Human analysis showed that the majority of the questions were relevant and useful, despite the entire dataset being synthetically generated. 

LegalBench.PT has significant potential for enhancement and expansion. Future work may focus on expanding underrepresented legal areas or developing tasks that assess capabilities beyond legal knowledge and reasoning, such as contract analysis or case summarization. Filtering out easy, ambiguous, or incorrect questions would also enhance the dataset's quality. Ultimately, thorough evaluation by human experts would ensure its reliability in assessing LLMs.

\section*{Limitations}

Although derived from real exams, LegalBench.PT's questions do not provide the same comprehensive assessment as the original exam questions. While the exams evaluate legal knowledge and its application to practical cases, they also assess reasoning and argumentation skills, which cannot be fully captured by multiple-choice and similar question formats. Additionally, as expected in a synthetically generated dataset, LegalBench.PT contains some noise, including questions with incorrect answers or improper legal terminology. Our analysis uncovered these issues, but since only a small sample was reviewed, these findings may not represent the entire benchmark. Human evaluation on a larger and more diverse set of questions validated the percentage of incorrect answers reported by the expert analysis, but also highlighted the presence of ambiguous questions, which can reduce the assessment accuracy.

\section*{Ethics Statement}
The creation of LegalBench.PT involved the use of real exam questions about the Portuguese legal system, which were synthetically rewritten into multiple-choice, true/false, and matching formats using GPT-4o. While we took steps to ensure the accuracy and relevance of the generated questions, including review by a legal professional, the dataset may still contain biases. These could stem from the synthetic generation process as well as inherent biases in the legal system itself. This benchmark is intended for research purposes and should not be used as a substitute for professional legal advice or decision-making. Furthermore, as LegalBench.PT focuses on the Portuguese legal system, care should be taken when generalizing the findings to other legal systems or jurisdictions. Users are encouraged to apply the dataset responsibly, keeping in mind its limitations and the specific legal context it covers.

\section*{Acknowledgments}

We would like to thank Equall for providing the resources for this work. We are also very grateful to all the lawyers who helped share this project or took the time to participate in the survey.

\bibliography{acl_latex}

\begin{thebibliography}{39}
\providecommand{\natexlab}[1]{#1}

\bibitem[{{Anthropic}(2024{\natexlab{a}})}]{claude-sonnet}
{Anthropic}. 2024{\natexlab{a}}.
\newblock \href {https://www.anthropic.com/news/claude-3-5-sonnet} {{Claude 3.5 Sonnet}}.
\newblock Accessed 30-Oct-2024.

\bibitem[{{Anthropic}(2024{\natexlab{b}})}]{claude}
{Anthropic}. 2024{\natexlab{b}}.
\newblock \href {https://www-cdn.anthropic.com/f2986af8d052f26236f6251da62d16172cfabd6e/claude-3-model-card.pdf} {{The Claude 3 Model Family: Opus, Sonnet, Haiku}}.

\bibitem[{Aumiller et~al.(2022)Aumiller, Chouhan, and Gertz}]{aumiller-etal-2022-eur}
Dennis Aumiller, Ashish Chouhan, and Michael Gertz. 2022.
\newblock \href {https://doi.org/10.18653/v1/2022.emnlp-main.519} {{EUR}-lex-sum: A multi- and cross-lingual dataset for long-form summarization in the legal domain}.
\newblock In \emph{Proceedings of the 2022 Conference on Empirical Methods in Natural Language Processing}, pages 7626--7639, Abu Dhabi, United Arab Emirates. Association for Computational Linguistics.

\bibitem[{Chalkidis et~al.(2019)Chalkidis, Androutsopoulos, and Aletras}]{chalkidis-etal-2019-neural}
Ilias Chalkidis, Ion Androutsopoulos, and Nikolaos Aletras. 2019.
\newblock \href {https://doi.org/10.18653/v1/P19-1424} {Neural legal judgment prediction in {E}nglish}.
\newblock In \emph{Proceedings of the 57th Annual Meeting of the Association for Computational Linguistics}, pages 4317--4323, Florence, Italy. Association for Computational Linguistics.

\bibitem[{Chalkidis et~al.(2021)Chalkidis, Fergadiotis, and Androutsopoulos}]{chalkidis-etal-2021-multieurlex}
Ilias Chalkidis, Manos Fergadiotis, and Ion Androutsopoulos. 2021.
\newblock \href {https://doi.org/10.18653/v1/2021.emnlp-main.559} {{M}ulti{EURLEX} - a multi-lingual and multi-label legal document classification dataset for zero-shot cross-lingual transfer}.
\newblock In \emph{Proceedings of the 2021 Conference on Empirical Methods in Natural Language Processing}, pages 6974--6996, Online and Punta Cana, Dominican Republic. Association for Computational Linguistics.

\bibitem[{Chalkidis et~al.(2022{\natexlab{a}})Chalkidis, Jana, Hartung, Bommarito, Androutsopoulos, Katz, and Aletras}]{chalkidis-etal-2022-lexglue}
Ilias Chalkidis, Abhik Jana, Dirk Hartung, Michael Bommarito, Ion Androutsopoulos, Daniel Katz, and Nikolaos Aletras. 2022{\natexlab{a}}.
\newblock \href {https://doi.org/10.18653/v1/2022.acl-long.297} {{L}ex{GLUE}: A benchmark dataset for legal language understanding in {E}nglish}.
\newblock In \emph{Proceedings of the 60th Annual Meeting of the Association for Computational Linguistics (Volume 1: Long Papers)}, pages 4310--4330, Dublin, Ireland. Association for Computational Linguistics.

\bibitem[{Chalkidis et~al.(2022{\natexlab{b}})Chalkidis, Pasini, Zhang, Tomada, Schwemer, and S{\o}gaard}]{chalkidis-etal-2022-fairlex}
Ilias Chalkidis, Tommaso Pasini, Sheng Zhang, Letizia Tomada, Sebastian Schwemer, and Anders S{\o}gaard. 2022{\natexlab{b}}.
\newblock \href {https://doi.org/10.18653/v1/2022.acl-long.301} {{F}air{L}ex: A multilingual benchmark for evaluating fairness in legal text processing}.
\newblock In \emph{Proceedings of the 60th Annual Meeting of the Association for Computational Linguistics (Volume 1: Long Papers)}, pages 4389--4406, Dublin, Ireland. Association for Computational Linguistics.

\bibitem[{Colombo et~al.(2024{\natexlab{a}})Colombo, Pires, Boudiaf, Melo, Culver, Morgado, Malaboeuf, Hautreux, Charpentier, and Desa}]{colombo2024saullm54bsaullm141bscaling}
Pierre Colombo, Telmo Pires, Malik Boudiaf, Rui Melo, Dominic Culver, Sofia Morgado, Etienne Malaboeuf, Gabriel Hautreux, Johanne Charpentier, and Michael Desa. 2024{\natexlab{a}}.
\newblock \href {https://arxiv.org/abs/2407.19584} {Saullm-54b \& saullm-141b: Scaling up domain adaptation for the legal domain}.
\newblock \emph{Preprint}, arXiv:2407.19584.

\bibitem[{Colombo et~al.(2024{\natexlab{b}})Colombo, Pires, Boudiaf, Culver, Melo, Corro, Martins, Esposito, Raposo, Morgado, and Desa}]{colombo2024saullm7bpioneeringlargelanguage}
Pierre Colombo, Telmo~Pessoa Pires, Malik Boudiaf, Dominic Culver, Rui Melo, Caio Corro, Andre F.~T. Martins, Fabrizio Esposito, Vera~Lúcia Raposo, Sofia Morgado, and Michael Desa. 2024{\natexlab{b}}.
\newblock \href {https://arxiv.org/abs/2403.03883} {Saullm-7b: A pioneering large language model for law}.
\newblock \emph{Preprint}, arXiv:2403.03883.

\bibitem[{Dai et~al.(2024)Dai, Feng, Huang, Jia, Xie, Zhang, Han, Tian, and Wang}]{dai2024laiwchineselegallarge}
Yongfu Dai, Duanyu Feng, Jimin Huang, Haochen Jia, Qianqian Xie, Yifang Zhang, Weiguang Han, Wei Tian, and Hao Wang. 2024.
\newblock \href {https://arxiv.org/abs/2310.05620} {Laiw: A chinese legal large language models benchmark}.
\newblock \emph{Preprint}, arXiv:2310.05620.

\bibitem[{Dubey et~al.(2024)Dubey, Jauhri, Pandey, Kadian, Al-Dahle, Letman, Mathur, Schelten, Yang, Fan et~al.}]{dubey2024llama3herdmodels}
Abhimanyu Dubey, Abhinav Jauhri, Abhinav Pandey, Abhishek Kadian, Ahmad Al-Dahle, Aiesha Letman, Akhil Mathur, Alan Schelten, Amy Yang, Angela Fan, et~al. 2024.
\newblock \href {https://arxiv.org/abs/2407.21783} {{The Llama 3 Herd of Models}}.
\newblock \emph{Preprint}, arXiv:2407.21783.

\bibitem[{Fei et~al.(2023)Fei, Shen, Zhu, Zhou, Han, Zhang, Chen, Shen, and Ge}]{fei2023lawbenchbenchmarkinglegalknowledge}
Zhiwei Fei, Xiaoyu Shen, Dawei Zhu, Fengzhe Zhou, Zhuo Han, Songyang Zhang, Kai Chen, Zongwen Shen, and Jidong Ge. 2023.
\newblock \href {https://arxiv.org/abs/2309.16289} {Lawbench: Benchmarking legal knowledge of large language models}.
\newblock \emph{Preprint}, arXiv:2309.16289.

\bibitem[{Guha et~al.(2023)Guha, Nyarko, Ho, Ré, Chilton, Narayana, Chohlas-Wood, Peters, Waldon, Rockmore, Zambrano, Talisman, Hoque, Surani, Fagan, Sarfaty, Dickinson, Porat, Hegland, Wu, Nudell, Niklaus, Nay, Choi, Tobia, Hagan, Ma, Livermore, Rasumov-Rahe, Holzenberger, Kolt, Henderson, Rehaag, Goel, Gao, Williams, Gandhi, Zur, Iyer, and Li}]{guha2023legalbenchcollaborativelybuiltbenchmark}
Neel Guha, Julian Nyarko, Daniel~E. Ho, Christopher Ré, Adam Chilton, Aditya Narayana, Alex Chohlas-Wood, Austin Peters, Brandon Waldon, Daniel~N. Rockmore, Diego Zambrano, Dmitry Talisman, Enam Hoque, Faiz Surani, Frank Fagan, Galit Sarfaty, Gregory~M. Dickinson, Haggai Porat, Jason Hegland, Jessica Wu, Joe Nudell, Joel Niklaus, John Nay, Jonathan~H. Choi, Kevin Tobia, Margaret Hagan, Megan Ma, Michael Livermore, Nikon Rasumov-Rahe, Nils Holzenberger, Noam Kolt, Peter Henderson, Sean Rehaag, Sharad Goel, Shang Gao, Spencer Williams, Sunny Gandhi, Tom Zur, Varun Iyer, and Zehua Li. 2023.
\newblock \href {https://arxiv.org/abs/2308.11462} {Legalbench: A collaboratively built benchmark for measuring legal reasoning in large language models}.
\newblock \emph{Preprint}, arXiv:2308.11462.

\bibitem[{Henderson et~al.(2022)Henderson, Krass, Zheng, Guha, Manning, Jurafsky, and Ho}]{NEURIPS2022_bc218a0c}
Peter Henderson, Mark Krass, Lucia Zheng, Neel Guha, Christopher~D Manning, Dan Jurafsky, and Daniel Ho. 2022.
\newblock \href {https://proceedings.neurips.cc/paper_files/paper/2022/file/bc218a0c656e49d4b086975a9c785f47-Paper-Datasets_and_Benchmarks.pdf} {Pile of law: Learning responsible data filtering from the law and a 256gb open-source legal dataset}.
\newblock In \emph{Advances in Neural Information Processing Systems}, volume~35, pages 29217--29234. Curran Associates, Inc.

\bibitem[{Hendrycks et~al.(2021{\natexlab{a}})Hendrycks, Burns, Basart, Zou, Mazeika, Song, and Steinhardt}]{hendrycks2021measuring}
Dan Hendrycks, Collin Burns, Steven Basart, Andy Zou, Mantas Mazeika, Dawn Song, and Jacob Steinhardt. 2021{\natexlab{a}}.
\newblock \href {https://openreview.net/forum?id=d7KBjmI3GmQ} {Measuring massive multitask language understanding}.
\newblock In \emph{International Conference on Learning Representations}.

\bibitem[{Hendrycks et~al.(2021{\natexlab{b}})Hendrycks, Burns, Chen, and Ball}]{NEURIPS_DATASETS_AND_BENCHMARKS2021_6ea9ab1b}
Dan Hendrycks, Collin Burns, Anya Chen, and Spencer Ball. 2021{\natexlab{b}}.
\newblock \href {https://datasets-benchmarks-proceedings.neurips.cc/paper_files/paper/2021/file/6ea9ab1baa0efb9e19094440c317e21b-Paper-round1.pdf} {Cuad: An expert-annotated nlp dataset for legal contract review}.
\newblock In \emph{Proceedings of the Neural Information Processing Systems Track on Datasets and Benchmarks}, volume~1.

\bibitem[{Hwang et~al.(2022)Hwang, Lee, Cho, Lee, and Seo}]{NEURIPS2022_d15abd14}
Wonseok Hwang, Dongjun Lee, Kyoungyeon Cho, Hanuhl Lee, and Minjoon Seo. 2022.
\newblock \href {https://proceedings.neurips.cc/paper_files/paper/2022/file/d15abd14d5894eebd185b756541d420e-Paper-Datasets_and_Benchmarks.pdf} {A multi-task benchmark for korean legal language understanding and judgement prediction}.
\newblock In \emph{Advances in Neural Information Processing Systems}, volume~35, pages 32537--32551. Curran Associates, Inc.

\bibitem[{Jiang et~al.(2024)Jiang, Sablayrolles, Roux, Mensch, Savary, Bamford, Chaplot, de~las Casas, Hanna, Bressand, Lengyel, Bour, Lample, Lavaud, Saulnier, Lachaux, Stock, Subramanian, Yang, Antoniak, Scao, Gervet, Lavril, Wang, Lacroix, and Sayed}]{jiang2024mixtralexperts}
Albert~Q. Jiang, Alexandre Sablayrolles, Antoine Roux, Arthur Mensch, Blanche Savary, Chris Bamford, Devendra~Singh Chaplot, Diego de~las Casas, Emma~Bou Hanna, Florian Bressand, Gianna Lengyel, Guillaume Bour, Guillaume Lample, Lélio~Renard Lavaud, Lucile Saulnier, Marie-Anne Lachaux, Pierre Stock, Sandeep Subramanian, Sophia Yang, Szymon Antoniak, Teven~Le Scao, Théophile Gervet, Thibaut Lavril, Thomas Wang, Timothée Lacroix, and William~El Sayed. 2024.
\newblock \href {https://arxiv.org/abs/2401.04088} {Mixtral of experts}.
\newblock \emph{Preprint}, arXiv:2401.04088.

\bibitem[{Joshi et~al.(2024)Joshi, Paul, Sharma, Goyal, Ghosh, and Modi}]{joshi2024ilturbenchmarkindianlegal}
Abhinav Joshi, Shounak Paul, Akshat Sharma, Pawan Goyal, Saptarshi Ghosh, and Ashutosh Modi. 2024.
\newblock \href {https://arxiv.org/abs/2407.05399} {Il-tur: Benchmark for indian legal text understanding and reasoning}.
\newblock \emph{Preprint}, arXiv:2407.05399.

\bibitem[{Junior et~al.(2024)Junior, Pires, Romero, and Nogueira}]{junior2024jurulegalbrazilianlarge}
Roseval~Malaquias Junior, Ramon Pires, Roseli Romero, and Rodrigo Nogueira. 2024.
\newblock \href {https://arxiv.org/abs/2403.18140} {Juru: Legal brazilian large language model from reputable sources}.
\newblock \emph{Preprint}, arXiv:2403.18140.

\bibitem[{Katz et~al.(2024)Katz, Bommarito, Gao, and Arredondo}]{doi:10.1098/rsta.2023.0254}
Daniel~Martin Katz, Michael~James Bommarito, Shang Gao, and Pablo Arredondo. 2024.
\newblock \href {https://doi.org/10.1098/rsta.2023.0254} {Gpt-4 passes the bar exam}.
\newblock \emph{Philosophical Transactions of the Royal Society A: Mathematical, Physical and Engineering Sciences}, 382(2270):20230254.

\bibitem[{Lopes et~al.(2024)Lopes, Magalhaes, and Semedo}]{lopes-etal-2024-gloria}
Ricardo Lopes, Joao Magalhaes, and David Semedo. 2024.
\newblock \href {https://aclanthology.org/2024.propor-1.45} {{G}l{\'o}r{IA}: A generative and open large language model for {P}ortuguese}.
\newblock In \emph{Proceedings of the 16th International Conference on Computational Processing of Portuguese - Vol. 1}, pages 441--453, Santiago de Compostela, Galicia/Spain. Association for Computational Lingustics.

\bibitem[{Malik et~al.(2021)Malik, Sanjay, Nigam, Ghosh, Guha, Bhattacharya, and Modi}]{malik-etal-2021-ildc}
Vijit Malik, Rishabh Sanjay, Shubham~Kumar Nigam, Kripabandhu Ghosh, Shouvik~Kumar Guha, Arnab Bhattacharya, and Ashutosh Modi. 2021.
\newblock \href {https://doi.org/10.18653/v1/2021.acl-long.313} {{ILDC} for {CJPE}: {I}ndian legal documents corpus for court judgment prediction and explanation}.
\newblock In \emph{Proceedings of the 59th Annual Meeting of the Association for Computational Linguistics and the 11th International Joint Conference on Natural Language Processing (Volume 1: Long Papers)}, pages 4046--4062, Online. Association for Computational Linguistics.

\bibitem[{Melo et~al.(2023)Melo, Santos, and Dias}]{10.1007/978-3-031-49011-8_12}
Rui Melo, Pedro~A. Santos, and Jo\~{a}o Dias. 2023.
\newblock \href {https://doi.org/10.1007/978-3-031-49011-8_12} {A semantic search system for the supremo tribunal de justi\c{c}a}.
\newblock In \emph{Progress in Artificial Intelligence: 22nd EPIA Conference on Artificial Intelligence, EPIA 2023, Faial Island, Azores, September 5–8, 2023, Proceedings, Part II}, page 142–154, Berlin, Heidelberg. Springer-Verlag.

\bibitem[{Niklaus et~al.(2023)Niklaus, Matoshi, Rani, Galassi, St{\"u}rmer, and Chalkidis}]{niklaus-etal-2023-lextreme}
Joel Niklaus, Veton Matoshi, Pooja Rani, Andrea Galassi, Matthias St{\"u}rmer, and Ilias Chalkidis. 2023.
\newblock \href {https://doi.org/10.18653/v1/2023.findings-emnlp.200} {{LEXTREME}: A multi-lingual and multi-task benchmark for the legal domain}.
\newblock In \emph{Findings of the Association for Computational Linguistics: EMNLP 2023}, pages 3016--3054, Singapore. Association for Computational Linguistics.

\bibitem[{Niklaus et~al.(2024{\natexlab{a}})Niklaus, Matoshi, St{\"u}rmer, Chalkidis, and Ho}]{niklaus-etal-2024-multilegalpile}
Joel Niklaus, Veton Matoshi, Matthias St{\"u}rmer, Ilias Chalkidis, and Daniel Ho. 2024{\natexlab{a}}.
\newblock \href {https://aclanthology.org/2024.acl-long.805} {{M}ulti{L}egal{P}ile: A 689{GB} multilingual legal corpus}.
\newblock In \emph{Proceedings of the 62nd Annual Meeting of the Association for Computational Linguistics (Volume 1: Long Papers)}, pages 15077--15094, Bangkok, Thailand. Association for Computational Linguistics.

\bibitem[{Niklaus et~al.(2024{\natexlab{b}})Niklaus, Zheng, McCarthy, Hahn, Rosen, Henderson, Ho, Honke, Liang, and Manning}]{niklaus2024flawnt5empiricalexaminationeffective}
Joel Niklaus, Lucia Zheng, Arya~D. McCarthy, Christopher Hahn, Brian~M. Rosen, Peter Henderson, Daniel~E. Ho, Garrett Honke, Percy Liang, and Christopher Manning. 2024{\natexlab{b}}.
\newblock \href {https://arxiv.org/abs/2404.02127} {Flawn-t5: An empirical examination of effective instruction-tuning data mixtures for legal reasoning}.
\newblock \emph{Preprint}, arXiv:2404.02127.

\bibitem[{OpenAI(2023)}]{openai2024gpt4technicalreport}
OpenAI. 2023.
\newblock \href {https://cdn.openai.com/papers/gpt-4.pdf} {Gpt-4 technical report}.

\bibitem[{{OpenAI}(2024{\natexlab{a}})}]{GPT-4o-mini-advancing-cost-efficient-intelligence}
{OpenAI}. 2024{\natexlab{a}}.
\newblock \href {https://openai.com/index/gpt-4o-mini-advancing-cost-efficient-intelligence/} {{GPT-4o mini: advancing cost-efficient intelligence}}.
\newblock Accessed 12-Oct-2024.

\bibitem[{{OpenAI}(2024{\natexlab{b}})}]{Hello-GPT-4o}
{OpenAI}. 2024{\natexlab{b}}.
\newblock \href {https://openai.com/index/hello-gpt-4o/} {{Hello GPT-4o}}.
\newblock Accessed 12-Oct-2024.

\bibitem[{Rodrigues et~al.(2023)Rodrigues, Gomes, Silva, Branco, Santos, Cardoso, and Osório}]{Rodrigues_2023}
João Rodrigues, Luís Gomes, João Silva, António Branco, Rodrigo Santos, Henrique~Lopes Cardoso, and Tomás Osório. 2023.
\newblock \href {https://doi.org/10.1007/978-3-031-49008-8_35} {\emph{Advancing Neural Encoding of Portuguese with Transformer Albertina PT-*}}, page 441–453.
\newblock Springer Nature Switzerland.

\bibitem[{Santos et~al.(2024)Santos, Silva, Gomes, Rodrigues, and Branco}]{santos-etal-2024-advancing}
Rodrigo Santos, Jo{\~a}o~Ricardo Silva, Lu{\'\i}s Gomes, Jo{\~a}o Rodrigues, and Ant{\'o}nio Branco. 2024.
\newblock \href {https://aclanthology.org/2024.sigul-1.3} {Advancing generative {AI} for {P}ortuguese with open decoder gerv{\'a}sio {PT}*}.
\newblock In \emph{Proceedings of the 3rd Annual Meeting of the Special Interest Group on Under-resourced Languages @ LREC-COLING 2024}, pages 16--26, Torino, Italia. ELRA and ICCL.

\bibitem[{Shen et~al.(2022)Shen, Lo, Yu, Dahlberg, Schlanger, and Downey}]{NEURIPS2022_552ef803}
Zejiang Shen, Kyle Lo, Lauren Yu, Nathan Dahlberg, Margo Schlanger, and Doug Downey. 2022.
\newblock \href {https://proceedings.neurips.cc/paper_files/paper/2022/file/552ef803bef9368c29e53c167de34b55-Paper-Datasets_and_Benchmarks.pdf} {Multi-lexsum: Real-world summaries of civil rights lawsuits at multiple granularities}.
\newblock In \emph{Advances in Neural Information Processing Systems}, volume~35, pages 13158--13173. Curran Associates, Inc.

\bibitem[{Stern et~al.(2024)Stern, Rasiah, Matoshi, Bose, Stürmer, Chalkidis, Ho, and Niklaus}]{stern2024lawlanguagesbenchmarkingmultilingual}
Ronja Stern, Vishvaksenan Rasiah, Veton Matoshi, Srinanda~Brügger Bose, Matthias Stürmer, Ilias Chalkidis, Daniel~E. Ho, and Joel Niklaus. 2024.
\newblock \href {https://arxiv.org/abs/2306.09237} {One law, many languages: Benchmarking multilingual legal reasoning for judicial support}.
\newblock \emph{Preprint}, arXiv:2306.09237.

\bibitem[{Wang et~al.(2023)Wang, Scardigli, Tang, Chen, Levkin, Chen, Ball, Woodside, Zhang, and Hendrycks}]{wang-etal-2023-maud}
Steven Wang, Antoine Scardigli, Leonard Tang, Wei Chen, Dmitry Levkin, Anya Chen, Spencer Ball, Thomas Woodside, Oliver Zhang, and Dan Hendrycks. 2023.
\newblock \href {https://doi.org/10.18653/v1/2023.emnlp-main.1019} {{MAUD}: An expert-annotated legal {NLP} dataset for merger agreement understanding}.
\newblock In \emph{Proceedings of the 2023 Conference on Empirical Methods in Natural Language Processing}, pages 16369--16382, Singapore. Association for Computational Linguistics.

\bibitem[{Yao et~al.(2022)Yao, Xiao, Wang, Liu, Hou, Tu, Li, Liu, Shen, and Sun}]{yao-etal-2022-leven}
Feng Yao, Chaojun Xiao, Xiaozhi Wang, Zhiyuan Liu, Lei Hou, Cunchao Tu, Juanzi Li, Yun Liu, Weixing Shen, and Maosong Sun. 2022.
\newblock \href {https://doi.org/10.18653/v1/2022.findings-acl.17} {{LEVEN}: A large-scale {C}hinese legal event detection dataset}.
\newblock In \emph{Findings of the Association for Computational Linguistics: ACL 2022}, pages 183--201, Dublin, Ireland. Association for Computational Linguistics.

\bibitem[{Zheng et~al.(2023)Zheng, Chiang, Sheng, Zhuang, Wu, Zhuang, Lin, Li, Li, Xing, Zhang, Gonzalez, and Stoica}]{NEURIPS2023_91f18a12}
Lianmin Zheng, Wei-Lin Chiang, Ying Sheng, Siyuan Zhuang, Zhanghao Wu, Yonghao Zhuang, Zi~Lin, Zhuohan Li, Dacheng Li, Eric Xing, Hao Zhang, Joseph~E Gonzalez, and Ion Stoica. 2023.
\newblock \href {https://proceedings.neurips.cc/paper_files/paper/2023/file/91f18a1287b398d378ef22505bf41832-Paper-Datasets_and_Benchmarks.pdf} {Judging llm-as-a-judge with mt-bench and chatbot arena}.
\newblock In \emph{Advances in Neural Information Processing Systems}, volume~36, pages 46595--46623. Curran Associates, Inc.

\bibitem[{Zhong et~al.(2020)Zhong, Xiao, Tu, Zhang, Liu, and Sun}]{Zhong_Xiao_Tu_Zhang_Liu_Sun_2020}
Haoxi Zhong, Chaojun Xiao, Cunchao Tu, Tianyang Zhang, Zhiyuan Liu, and Maosong Sun. 2020.
\newblock \href {https://doi.org/10.1609/aaai.v34i05.6519} {Jec-qa: A legal-domain question answering dataset}.
\newblock \emph{Proceedings of the AAAI Conference on Artificial Intelligence}, 34(05):9701--9708.

\bibitem[{Zhou et~al.(2024)Zhou, Shi, Song, Yang, Jin, Guo, and Li}]{zhou2024lawgptchineselegalknowledgeenhanced}
Zhi Zhou, Jiang-Xin Shi, Peng-Xiao Song, Xiao-Wen Yang, Yi-Xuan Jin, Lan-Zhe Guo, and Yu-Feng Li. 2024.
\newblock \href {https://arxiv.org/abs/2406.04614} {Lawgpt: A chinese legal knowledge-enhanced large language model}.
\newblock \emph{Preprint}, arXiv:2406.04614.

\end{thebibliography}

\newpage
\appendix


\section{Question Generation Details}
\label{sec:question-generation-details}

We recall the three approaches implemented for generating new questions:
\begin{enumerate}
    \item Providing the model with a group of short, independent question-answer pairs.
    \item Feeding the model one question-answer pair at a time. These questions usually include a long problem statement.
    \item Presenting the model with a set of exam questions and answers related to a common problem statement. These exam questions sometimes present new assumptions, as a continuation of the problem statement, that may contain critical information.
\end{enumerate}

\begin{table*}[ht!]
\scriptsize
\centering
\begin{tabular}{p{0.45\textwidth}|p{0.45\textwidth}} \toprule
\multicolumn{1}{c}{\textbf{Exam Questions}} & \multicolumn{1}{c}{\textbf{Output}} \\ \midrule
\parbox[t]{\linewidth}{
What distinguishes a federation from a real union? \\

A federation is a composite state, made up of several federated states, which, despite having their own Constitution, do not enjoy sovereignty, being subordinate to the federal Constitution.
The real union, in turn, although also corresponding to a composite state, differs from the federation because the states that comprise it retain their autonomy and adopt a common Constitution, along with common bodies.
Thus, while in the federation there is an overlap, in the real union there is a fusion. \\
\\ 
\textbf{====================} \\
\\
How can it be explained that, in the United States of America, a candidate can be elected President even though they obtained fewer votes than their opponent? \\
\\
Topic to develop: this possibility – which, moreover, has occurred more than once – is explained by the peculiar presidential election system that, for historical reasons, was enshrined in the United States of America and which, for various reasons (including the difficulty of amending the Constitution), is still in force today: the indirect suffrage system.
Since the President is elected by an electoral college, which, in turn, is composed of representatives elected by each federated state (the number of representatives per state being larger the greater its population), through a majority electoral system, this means that, more than the total number of votes obtained by each candidate nationwide, what matters most for the outcome of the presidential elections is the number of votes obtained by each candidate in the states that elect a larger number of representatives to the electoral college.  \\
\\
\textbf{====================} \\
\\
\textit{Other questions omitted} \\ }
&
\parbox[t]{\linewidth}{
\textbf{Type:} Multiple Choice \\
\textbf{Question:} What is the main difference between a federation and a real union? \\
a) In the federation, the federated states maintain their sovereignty. \\
b) In the real union, the states retain their autonomy and adopt a common Constitution. \\
c) In the federation, there is a fusion of states. \\
d) In the real union, the states do not have autonomy. \\
\textbf{Answer:} b) In the real union, the states retain their autonomy and adopt a common Constitution. \\
\\
\textbf{Type:} Case Analysis Questions \\
\textbf{Question:} A presidential candidate in the United States of America obtained fewer votes than their opponent but was elected President. What is the correct explanation for this situation? \\
a) The U.S. electoral system is based on direct suffrage. \\
b) The President is elected by an electoral college, composed of representatives elected by each state. \\
c) A candidate with fewer votes cannot be elected President. \\
d) The U.S. electoral system is based on the proportionality of votes. \\
\textbf{Answer:} b) The President is elected by an electoral college, composed of representatives elected by each state. \\
\\
\textit{Other questions omitted} \\ } \\
\bottomrule
\end{tabular}
\caption{Constitutional Law: English-translated example of exam question-answer pairs fed into GPT-4o and part of the respective output.} \label{table:exam-prompt-sem}
\end{table*}

\begin{table*}[ht!]
\scriptsize
\centering
\begin{tabular}{p{0.45\textwidth}|p{0.45\textwidth}} \toprule
\multicolumn{1}{c}{\textbf{Exam Question}} & \multicolumn{1}{c}{\textbf{Output}} \\ \midrule
\parbox[t]{\linewidth}{
On February 2, 2022, Artur met Benedita at the Vilamoura casino, and he fell in love with her at first sight. Benedita, who has a gambling addiction, ended up owing €80,000.00 to the Algarvian mafia, which now threatens to kill her if she doesn’t pay her debt.
Taking advantage of the situation, Artur promises to pay the €80,000.00, sparing Benedita from the mafia’s retaliation, if she marries him by Valentine's Day. Desperate, Benedita agrees to the request and marries Artur without a preliminary process, due to the urgency. Eight months later, Benedita discovers that Artur is a criminal wanted in Italy for counterfeiting Gucci bags.
Assess the validity of the marriage between Artur and Benedita, considering all the described facts. (5 points) \\
\\
The factual matter primarily relates to the causes of marriage annulment.
First of all, since the marriage was entered into under the promise of paying Benedita’s €80,000.00 debt to the mafia, in order to spare her from retaliation, this is a cause for annulment due to moral coercion (Article 1631(b), 1638(2), and Article 282(1)). However, for the marriage to be annulled on this basis of coercion, it will be necessary to file for annulment (Article 1632) by the persons indicated in Article 1641, within the time frame established by Article 1645. In this case, it seems that the right to annul the marriage on the grounds of moral coercion may have already expired.
On the other hand, the fact that Artur is a criminal wanted in Italy may be relevant for invoking annulment based on an error of will (Articles 1631(b) and 1636). For the error of will to be valid under the mentioned terms, the following requirements must be met: i) an error concerning essential qualities of the person; ii) an excusable error; iii) both objective and subjective essentiality. In this case, the error is relevant, as it concerns essential qualities of the other spouse (being a criminal); it is excusable (not a gross error), since Artur is wanted in Italy and not in Portugal, nor is it necessarily a notorious fact, so it is excusable that Benedita did not know;  \textit{[continues...] \\ }
}
&
\parbox[t]{\linewidth}{
\textbf{Context:} \\
On February 2, 2022, Artur met Benedita at the Vilamoura casino, and he fell in love with her at first sight. Benedita, who has a gambling addiction, ended up owing €80,000.00 to the Algarvian mafia, which now threatens to kill her if she doesn’t pay her debt.
Taking advantage of the situation, Artur promises to pay the €80,000.00, sparing Benedita from the mafia’s retaliation, if she marries him by Valentine's Day. Desperate, Benedita agrees to the request and marries Artur without a preliminary process, due to the urgency. Eight months later, Benedita discovers that Artur is a criminal wanted in Italy for counterfeiting Gucci bags. \\
\\
\textbf{Type:} Multiple Choice \\
\textbf{Question:} Which of the following options is a cause for the annulment of the marriage between Artur and Benedita? \\
a) Lack of true love \\
b) Moral coercion \\
c) Age difference \\
d) Lack of witnesses \\
\textbf{Answer:} b \\
\\
\textbf{Type:} Multiple Selection Questions \\
\textbf{Question:} Which of the following factors can be considered moral coercion in the context of marriage? \\
a) Death threats \\
b) Financial pressure \\
c) Emotional blackmail \\
d) Age difference \\
\textbf{Answer:} a) b) c) \\
\\
\textit{Other questions omitted} \\ } \\
\bottomrule
\end{tabular}
\caption{Family Law: English-translated example of exam question-answer pair fed into GPT-4o and part of the respective output.} \label{table:exam-prompt-meio}
\end{table*}

\begin{table*}[ht]
\scriptsize
\centering
\begin{tabular}{p{0.45\textwidth}|p{0.45\textwidth}} \toprule
\multicolumn{1}{c}{\textbf{Exam Questions}} & \multicolumn{1}{c}{\textbf{Output}} \\ \midrule
\parbox[t]{\linewidth}{
Adalberto has been running a vegetarian restaurant in Chiado for several years. Tired of the daily grind, he decides to contact Berta to find someone interested in taking over the business. Berta quickly starts working, speaking wonders about the restaurant and convincing Carloto to take over the business.  
On the agreed date, Adalberto and Carloto sign a contract that states: “the restaurant is sold with all its elements, including the exclusive supply contract with Dioclécio.”  
Two months later, during a friendly conversation, Carloto convinces Ermelinda to invest in his business in exchange for a share of the profits he may obtain. 
Answer the following questions, justifying your response: \\
\\
1. Berta demands compensation from Adalberto for her services. Adalberto claims that, aside from no agreement being made regarding this, he also owes her nothing because Carloto hasn’t paid for the business. Quid iuris? (4 marks) \\
\\
Classification of the contract between Adalberto and Berta as a mediation contract. Explanation of the applicable legal regime, considering the absence of a general regime for this type of contract.  
Distinction between the mediation contract and other contractual types, particularly agency contracts.  
Analysis of the right to remuneration: in mediation, unlike in agency contracts, the commission is due regardless of payment by the buyer, as long as the contract is signed. The lack of an agreement between the parties on the amount of remuneration does not prevent payment.  \\
\\
2. Dioclécio is contacted by Carloto, placing several product orders. Dioclécio is surprised because, aside from not knowing who Carloto is, he never agreed to anything with him. Moreover, the commercial conditions agreed with Adalberto stemmed from their long-standing friendship. Quid iuris? (4 marks) \\
\\
Analysis of the situation at hand in light of the effects of business transfer on contracts related to the establishment.  
Explanation of the key issues and various doctrinal positions, particularly: (i) application of the general regime for assignment of contractual position (Articles 424 and following), (ii) operational legal situations; (iii) existence of a general principle (derived from specific normative examples in Commercial Law) of automatic transfer of contractual positions related to the business establishment; \textit{[continued...]} \\ }
&
\parbox[t]{\linewidth}{
\textbf{Context:} \\
Adalberto has been running a vegetarian restaurant in Chiado for several years. Tired of the daily grind, he decides to contact Berta to find someone interested in taking over the business. Berta quickly starts working, speaking wonders about the restaurant and convincing Carloto to take over the business.
On the agreed date, Adalberto and Carloto sign a contract that states: “the restaurant is sold with all its elements, including the exclusive supply contract with Dioclécio.”
Two months later, during a friendly conversation, Carloto convinces Ermelinda to invest in his business in exchange for a share of the profits he may obtain. \\
\\
\textbf{Assumption:} \\
Berta demands compensation from Adalberto for her services. Adalberto claims that, aside from no agreement being made regarding this, he also owes her nothing because Carloto hasn’t paid for the business. \\

\textbf{Type:} True/False \\
\textbf{Question:} The lack of an agreement between the parties on the amount of remuneration prevents the payment of the commission in a mediation contract. \\
\textbf{Answer:} False \\
\\
\textit{Other questions omitted} \\
\\
\textbf{Assumption:} \\
Dioclécio is contacted by Carloto, placing several product orders. Dioclécio is surprised because, aside from not knowing who Carloto is, he never agreed to anything with him. Moreover, the commercial conditions agreed with Adalberto stemmed from their long-standing friendship. \\
\\
\textbf{Type:} Cloze Tasks \\
\textbf{Question:} The automatic transfer of contractual positions related to the business establishment is limited when: \\
a) The assignee offers greater guarantees than the assignor \\
b) The contract is \textit{intuitu personae} \\
c) The contract is for the supply of products \\
d) The contract is for the provision of services \\
\textbf{Answer:} b) The contract is \textit{intuitu personae} \\
\\
\textit{Other questions omitted} \\ } \\
\bottomrule
\end{tabular}
\caption{Commercial Law: English-translated example of exam question-answer pairs fed into GPT-4o and part of the respective output.} \label{table:exam-prompt-com}
\end{table*}

\begin{table*}[ht!]
\scriptsize
\centering
\begin{tabular}{p{0.95\linewidth}} 
\toprule
\textbf{Statement:} \\
GELL\&CO and SOLIMPA, two companies in the cleaning products manufacturing sector, decided in May 2020 to adapt their production to manufacturing hand sanitizer. On 02/08/2020, both companies were contacted by a representative of a well-known international distributor, JACQUES SILVA, from a Spanish branch, requesting the urgent delivery of 5,000 units of hand sanitizer. GELL\&CO and SOLIMPA accepted (each thinking it had been the only one contacted and contracted). GELL\&CO shipped merchandise valued at 15,000 euros to the address provided by JACQUES (located in Spain) on 05/09/2020. SOLIMPA shipped its merchandise, also valued at 15,000 euros, on 10/09/2020. At the end of September 2020, having received no payment and unable to contact JACQUES, the companies individually contacted the international distributor, confirming that there was no representative named JACQUES, nor any contract. On 06/10/2020, the Portuguese companies filed a complaint with the Public Prosecutor's Office. \\
\\
\textbf{Assumption:} \\
On 15/03/2021, following an undercover operation, the Judiciary Police (PJ) arrested MÁRIO MENDES, suspected of impersonating JACQUES SILVA, and the main suspect in several fraud crimes in both Portugal and Spain, in Elvas. On 16/03/2021, MÁRIO was brought before the Criminal Investigation Judge (JIC) for his first interrogation as a detained defendant. \\
\\
{[Some questions related to the assumption above, to which you have already answered, were omitted.]} \\
\\
\textbf{Assumption:} \\
On 20/09/2021, the Public Prosecutor's Office charged MÁRIO with two counts of aggravated fraud, in actual competition (Article 218/1 of the Criminal Code), one against GELL\&CO and another against SOLIMPA, as well as one count of identity document forgery (Article 256/1 of the Criminal Code), also in actual competition. \\
\\
{[Some questions related to the assumption above, to which you have already answered, were omitted.]} \\
\\
\textbf{Now consider the following assumption:} \\
Before the trial began, MÁRIO reached an agreement with GELL\&CO and SOLIMPA, voluntarily compensating for the damages caused, in accordance with Articles 218/4 and 206/1 of the Criminal Code. Consequently, the Criminal Investigation Judge declared MÁRIO’s criminal liability for the two counts of aggravated fraud extinguished, leaving the case to proceed to trial solely for the crime of document forgery. In his defense, MÁRIO provided evidence that he had used a false name in emails but had never fabricated or used a false document, as he had not used any symbols or identifying words of any registered trademarks. However, the trial court convicted him, invoking the agreement under Article 206 as an implicit confession of the commission of all crimes listed in the indictment. \\
\bottomrule
\end{tabular}
\caption{\label{table:example-all-assumptions-together}
Criminal Procedure Law example: illustration of how we combine a problem statement with three assumptions, all extracted from a GPT-4's output. The text presented is placed at the beginning of all questions generated by GPT-4o that follow the third assumption.}
\end{table*}

Tables~\ref{table:exam-prompt-sem}, \ref{table:exam-prompt-meio}, and \ref{table:exam-prompt-com} show English-translated examples of exam questions fed into GPT-4o, along with part of the corresponding outputs, illustrating the different approaches used and the output templates we requested.

For approach~\ref{question-generation-third-approach}, we simply requested the outputs to be sequences of tuples, with each tuple identifying the generated question along with its type and answer, as demonstrated in Table~\ref{table:exam-prompt-sem}. Table~\ref{table:exam-prompt-meio} illustrates a similar template that we requested for approach~\ref{question-generation-second-approach}. Since these exam questions usually present long problem statements, namely case analysis or excerpts from court decisions, we instruct the model to first identify the statement and then present the tuples with the new generated questions. Finally, for approach~\ref{question-generation-first-approach}, we adopted a slightly different template displayed in Table~\ref{table:exam-prompt-com}. In this case, the exam questions sometimes present new assumptions, as a continuation of the problem statement, that may contain critical information. We request the model to identify the problem statement, the assumptions, and the tuples with the new questions.

To get the final version of the generated questions: for approach~\ref{question-generation-second-approach}, we joined at the beginning of each question the respective statement; for approach~\ref{question-generation-first-approach}, we joined at the beginning of each question the respective statement and all assumptions preceding the question in the output. This method allows us to avoid losing important information introduced in previous assumptions.

Table~\ref{table:example-all-assumptions-together} illustrates an example of how we join a statement together with some assumptions. In this example, the dependence on the first assumption to understand the second and the third is clear. For instance, Mário Mendes and his role in the story are presented in the first assumption, a fact that is important to understand the following ones. Because this dependence between assumptions might not always apply, and, in some cases, assumptions may even be contradictory, we follow each assumption, except the last one, by the Portuguese version of ``Some questions related to the assumption above, to which you have already answered, were omitted.'' We did this to mimic a human taking the original exam, who would see all the questions together and have access to all the available information. However, we wanted to avoid the model conditioning its answers on previous questions and responses.

Table~\ref{table:prompt-english-generation} shows the English translation of a prompt template used to generate new questions from groups of short and independent exam questions (approach~\ref{question-generation-third-approach}, first iteration).

\begin{table}[ht!]
\scriptsize
\centering
\begin{tabular} {p{0.95\linewidth}} 
\toprule
Below, you are presented with a set of questions taken from exams and their respective grading criteria.
Your task is to create a new exam to evaluate a group of students, based on the questions and grading
criteria presented in 'Questions taken from exams'. You should create
various types of questions: Multiple Choice questions, Cloze Tasks,
True/False questions, Multiple Selection Questions,
Case Analysis Questions where a brief scenario/case is presented
and students are asked to choose the best answer from several options. Be creative. The purpose of the questions you will create is to assess how well a
student masters the topic {}. The questions you create should be challenging. The answers to the questions you create should be contained in 'Questions taken from exams'.
Avoid creating questions about specific articles or laws. Students are not required to memorize articles and laws and will not have access to them during the exam.
The questions you create should focus on assessing legal reasoning.
Create at least three to four questions of each of the types described above. The questions you create should all be independent of each other. \\
\\ 
Your output should follow the following format: \\
\\ 
'Type: \{\{insert question type here\}\} \\
Question: \{\{insert question here\}\} \\
Answer: \{\{insert answer here\}\} \\
\\ 
Type: \{\{insert question type here\}\} \\
Question: \{\{insert question here\}\} \\
Answer: \{\{insert answer here\}\} \\
\\ 
...' \\
\\ 
Attention: \\
i) For Cloze Tasks, present answer options, multiple-choice style - the answer should be only ONE letter, corresponding to the correct option; \\
ii) In True/False questions, the answer should be only 'True' or 'False'; \\
iii) In Multiple Selection Questions, present the answers in the format 'letter) letter) ...'; \\
iv) In Multiple Choice and Case Analysis Questions, the answer should be only ONE letter, corresponding to the correct option. \\
\\ 
Questions taken from exams: \\
'\{\}' \\ \bottomrule
\end{tabular}
\caption{\label{table:prompt-english-generation}
English translation of a prompt template used to generate new question-answer pairs.}
\end{table}

\section{Filtering Repeated Questions}
\label{sec:filtering-repeated-questions}

Table~\ref{table:processing-matching-questions} illustrates the processing applied to matching questions for some comparisons: we removed the first line (usually ``Match the items...'') and the letters/numbers identifying options.

\begin{table}[ht!]
\scriptsize
\centering
\begin{tabular}{p{6.5cm}} \toprule
{\textbf{Original matching question}} \\ \midrule
Match each item on the agenda with the type of competence involved.\\
a) Approval of the Code of Ethics and Conduct of the Municipality of Amadora\\
b) Acceptance of donations of works of art\\
c) Approval of the amendment to the inter-administrative contract for delegation of powers with the Parish of Alfragide\\
\\
1) Competence of the City Council\\
2) Competence of the Municipal Assembly\\
3) Competence of the Parish Council\\
\midrule
{\textbf{Processed matching question}} \\ \midrule
Approval of the Code of Ethics and Conduct of the Municipality of Amadora\\
Acceptance of donations of works of art\\
Approval of the amendment to the inter-administrative contract for delegation of powers with the Parish of Alfragide\\
Competence of the City Council\\
Competence of the Municipal Assembly\\
Competence of the Parish Council\\ \bottomrule
\end{tabular}
\caption{\label{table:processing-matching-questions}
Illustration of the processing applied to matching questions for lexical comparisons using ROUGE-L at summary level.
}
\end{table}

We recall that multiple-choice, cloze tasks, case analysis, and multiple selection questions are considered multiple-choice variants. For the lexical comparisons, Table~\ref{tab:summary-rougeL-within} and Table~\ref{tab:summary-rougeL-between} summarize the ROUGE-L variants, question processing, and thresholds used for filtering within and between the different question types. For the semantic comparisons, we considered two questions repeated or similar if the cosine similarity of their sentence embeddings was $0.80$ or higher. Table~\ref{tab:summary-semantic-within-between} summarizes how questions were processed based on their types. We manually set the best threshold for each scenario. 

\begin{table*}[ht!]
\footnotesize
\centering
\begin{tabular}{|c|c|c|c|c|}
\hline
\textbf{Type} & \textbf{Processing} & \multirow{2}{*}
{\textbf{\begin{tabular}[c]{@{}c@{}}ROUGE-L\\ Variant\end{tabular}}} & \textbf{Threshold} \\ 
 & & & \\ \hline
Matching questions & Illustrated in Table~\ref{table:processing-matching-questions} & Summary & $0.70$ \\ \hline
Multiple-choice variants & None & Summary & $0.80$ \\ \hline
Multiple-choice variants & Remove options & \multirow{2}{*}{Sentence} & \multirow{2}{*}{$0.75$} \\
\cline{1-2} True/False & None &  &  \\ \hline
\end{tabular}
\caption{\label{tab:summary-rougeL-within}
Summary of ROUGE-L comparisons within each question type (multiple-choice, cloze tasks, case analysis, multiple selection, true/false, matching questions — the first four referred to as multiple-choice variants): question processing, ROUGE-L variant used, and threshold applied.}
\end{table*}

\begin{table*}[ht!]
\footnotesize
\centering
\begin{tabular}{|c|c|c|c|}
\hline
\textbf{Type I} & \textbf{Type II} & \multirow{2}{*}
{\textbf{\begin{tabular}[c]{@{}c@{}}ROUGE-L\\ Variant\end{tabular}}} & \textbf{Threshold} \\ 
 & & & \\ \hline
\multirow{2}{*}{\begin{tabular}[c]{@{}c@{}}Multiple-choice variant\\ (remove options)\end{tabular}} & \begin{tabular}[c]{@{}c@{}}Multiple-choice variant\\ (remove options)\end{tabular} & \multirow{2}{*}{Sentence} & \multirow{2}{*}{0.85} \\ \cline{2-2} & \begin{tabular}[c]{@{}c@{}}True/False\\ (no processing)\end{tabular}  & & \\ \hline
\multirow{2}{*}{\begin{tabular}[c]{@{}c@{}}Multiple-choice variant\\ (no processing)\end{tabular}} & \begin{tabular}[c]{@{}c@{}}Multiple-choice variant\\ (no processing)\end{tabular}  & \multirow{2}{*}{Summary}  & \multirow{2}{*}{0.80} \\ \cline{2-2} & \begin{tabular}[c]{@{}c@{}}True/False\\ (no processing)\end{tabular} & & \\ \hline
\end{tabular}
\caption{\label{tab:summary-rougeL-between}
Summary of ROUGE-L comparisons between questions of different types (multiple-choice, cloze tasks, case analysis, multiple selection, true/false — the first four referred to as multiple-choice variants): question processing, ROUGE-L variant used, and threshold applied. Type I and Type II represent distinct types.}
\end{table*}

\begin{table}[ht!]
\footnotesize
\centering
\begin{tabular}{|c|c|l}
\cline{1-2}
\textbf{Type I} & \textbf{Type II} &  \\ \cline{1-2}
\begin{tabular}[c]{@{}c@{}}Matching question\\ (no processing)\end{tabular} & \begin{tabular}[c]{@{}c@{}}Any type\\ (no processing)\end{tabular} & \\ \cline{1-2}
\begin{tabular}[c]{@{}c@{}}Multiple-choice variant\\ (remove options)\end{tabular}  & \begin{tabular}[c]{@{}c@{}}Multiple-choice variant\\ (remove options)\end{tabular} & \\ \cline{2-2} & \begin{tabular}[c]{@{}c@{}}True/False\\ (no processing)\end{tabular} & \\ \cline{1-2}
\end{tabular}
\caption{Summary of semantic comparisons: compared types and processing. Type I and Type II may be the same.}
\label{tab:summary-semantic-within-between}
\end{table}

\section{More Statistics}
\label{sec:more-statistics}

Table~\ref{table:types-classes} shows the number of questions in the benchmark by type, and Table~\ref{table:FDSmodels-results-categories} presents the distribution of questions within each area of law across the different types, in percentage. The variation in the number of questions across the different types, as shown in Table~\ref{table:types-classes}, as well as the differences in distribution across areas, are partly due to the number of exam exercises run with each generation approach. In all approaches, we instructed GPT-4o to generate multiple-choice, cloze tasks, true/false, and multiple selection questions. In the approach using short, independent exam questions, we also instructed the model to generate case analysis questions, as the exam questions were often very theoretical. Conversely, matching questions were generated only in the other two approaches. Some variations in distribution across areas may stem from smaller subsets of questions, as in the case of Labor Procedure Law. Others differences are likely attributed to the filtering phase.

\begin{table}[ht]
\centering
\footnotesize
\begin{tabular}{lcc}
\toprule
\textbf{Type} & \textbf{Questions} \\ \midrule
Multiple-choice & 1099 \\
Cloze tasks & 983 \\
Case analysis & 89 \\
True/False & 1309 \\
Multiple selection & 695 \\
Matching questions & 548 \\ \bottomrule
\end{tabular}
\caption{\label{table:types-classes}
LegalBench.PT: number of questions by type.}
\end{table}

\begin{table*}[ht!]
\scriptsize
\centering
\begin{tabular}{lcccccc}
\toprule
\textbf{Fields of Law} & \textbf{Multiple-choice} & \textbf{Cloze tasks} & \textbf{Case analysis} & \textbf{True/False} & \textbf{Multiple selection} & \textbf{Matching questions} \\
\midrule
\textbf{Public} & 22.5 & 21.5 & 1.6 & 28.7 & 14.3 & 11.5 \\ 
\midrule
Environmental Law & 23.1 & 26.9 & 5.8 & 25.0 & 11.5 & 7.7 \\
Administrative Law & 22.5 & 22.9 & 2.2 & 27.8 & 12.8 & 11.9 \\
Constitutional Law & 22.8 & 22.8 & 3.9 & 26.2 & 15.5 & 8.7 \\
Energy Law & 9.5 & 14.3 & 19.0 & 33.3 & 23.8 & 0 \\
Public Finance Law & 20.2 & 21.6 & 2.4 & 29.3 & 13.5 & 13.0 \\
Financial Law & 25.0 & 31.2 & 0 & 25.0 & 0 & 18.8 \\
Tax Law & 21.1 & 22.4 & 0 & 28.9 & 15.9 & 11.6 \\
Criminal Law & 21.5 & 22.8 & 0.3 & 27.7 & 14.4 & 13.3 \\
Administrative Procedure Law & 21.6 & 15.5 & 1.7 & 31.9 & 19.0 & 10.3 \\
Civil Procedure Law & 22.4 & 21.7 & 0.7 & 29.0 & 13.6 & 12.5 \\
Criminal Procedure Law & 27.0 & 17.9 & 0.7 & 29.6 & 12.4 & 12.4 \\
Labor Procedure Law & 0 & 33.3 & 33.3 & 33.3 & 0 & 0 \\ 
Urban Planning Law & 24.5 & 21.4 & 0 & 31.6 & 16.3 & 6.1 \\
\midrule
\textbf{Private} & 23.6 & 20.5 & 1.8 & 26.6 & 14.9 & 12.5 \\
\midrule
Contract Law & 24.1 & 19.8 & 1.4 & 26.9 & 15.1 & 12.7 \\
Family Law & 23.6 & 17.3 & 4.1 & 29.1 & 14.5 & 11.4 \\
Law of Obligations & 28.4 & 19.7 & 0.5 & 26.1 & 12.4 & 12.8 \\
Property Law & 23.5 & 22.1 & 0.4 & 25.1 & 15.7 & 13.1 \\
Succession Law & 29.4 & 11.8 & 0 & 29.4 & 5.9 & 23.5 \\
Commercial Law & 22.9 & 19.6 & 1.7 & 26.7 & 15.0 & 14.2 \\
Banking Law & 16.7 & 30.0 & 0 & 26.7 & 16.7 & 10.0 \\
Maritime Law & 25.0 & 12.5 & 25.0 & 25.0 & 12.5 & 0 \\
Corporate Law & 24.7 & 18.2 & 0.6 & 29.2 & 15.6 & 11.7 \\
Securities Law & 20.0 & 22.2 & 4.4 & 22.2 & 20.0 & 11.1 \\ 
Transportation Law & 26.9 & 23.1 & 11.5 & 19.2 & 19.2 & 0 \\
Aviation Law & 22.6 & 25.8 & 12.9 & 22.6 & 16.1 & 0 \\
Insolvency Law & 27.0 & 16.2 & 0 & 29.7 & 16.2 & 10.8 \\
Private International Law & 23.6 & 22.7 & 1.8 & 28.2 & 11.8 & 11.8 \\
Labor Law & 18.4 & 22.2 & 2.4 & 27.1 & 15.0 & 15.0 \\
\midrule
\textbf{Public-Private} (Competition Law) & 24.6 & 21.1 & 0 & 25.1 & 18.3 & 10.9 \\
\midrule
\textbf{Public International} & 26.9 & 16.0 & 6.1 & 28.8 & 16.0 & 6.1 \\
\midrule
\textbf{EU and Community} & 23.8 & 21.3 & 3.0 & 30.5 & 12.8 & 8.5 \\
\midrule
\textbf{Overall} & 23.3 & 20.8 & 1.9 & 27.7 & 14.7 & 11.6 \\
\bottomrule
\end{tabular}
\caption{\label{table:FDSmodels-results-categories}
Distribution ($\%$) of questions within each area of law across the different types. Each row sums to $100\%$.}
\end{table*}

\section{Evaluation Prompts}
\label{sec:prompts-evaluation}

Table~\ref{table:example-evaluation-prompts} illustrates the English translation of an evaluation prompt. The ``Instruction'' field and the final line specifying the output format vary depending on the question type. The ``Statement'' and ``Assumption'' fields are omitted if a question does not have any statement or assumption associated.

\begin{table*}[ht!]
\scriptsize
\centering
\begin{tabular}{p{0.95\linewidth}} 
\toprule
You are solving an exam on Family Law. \\
\\
Statement: \\
Nuno and Sofia, both single and without children, got married on October 15, 2023, having previously celebrated, on January 15, 2023, a prenuptial agreement, in which they stipulated the following:
a) they adopt the regime of acquired community property, but all assets acquired with the couple's money, even in part (even if minimal), are common assets;
b) each spouse may alienate their own property without the need for the other spouse's consent;
c) the couple's assets are liable only for debts incurred by both spouses. \\
\\
Assumption: \\
Nuno and Sofia became parents to João on November 15, 2023. \\
\\
Instruction: \\
Answer the following multiple-choice question. Only one option is correct. \\
\\
Question: \\
The paternity of João is, in principle, established by \_\_\_\_\_\_. \\
a) declaration of paternity\\
b) presumption of paternity \\
c) judicial recognition \\
d) acknowledgment \\
\\
The output should only be: ``The correct answer is: \{letter\}'' \\
\bottomrule
\end{tabular}
\caption{\label{table:example-evaluation-prompts}
English-translated example of an evaluation prompt.}
\end{table*}

\section{Model Performance Across Types}
\label{sec:appendix-models-performance}

\begin{table*}[ht!]
\centering
\scriptsize
\begin{tabular}{lcccccc}
\toprule
\textbf{Models} & \textbf{Multiple-choice} & \textbf{Cloze tasks} & \textbf{Case analysis} & \textbf{True/False} & \textbf{Multiple selection} & \textbf{Matching questions} \\ \midrule
GPT-4o & 84.6 & 83.6 & 84.6 & 83.3 & 88.6 & 91.6 \\
GPT-4o-mini & 78.0 & 78.7 & 81.6 & 76.6 & 85.7 & 88.3 \\
Claude-3-Opus & 83.9 & 86.0 & 84.6 & 79.2 & 87.0 & 91.3 \\
Claude-3.5-Sonnet & 84.7 & 84.9 & 83.7 & 81.8 & 87.9 & 91.0 \\
Llama-3.1-8B & 69.2 & 68.3 & 72.7 & 64.4 & 78.4 & 64.8 \\
Llama-3.1-70B & 80.5 & 80.2 & 85.5 & 76.9 & 84.4 & 85.6 \\
Llama-3.1-405B & 85.2 & 82.4 & 81.9 & 80.6 & 85.6 & 89.4 \\
Mixtral-8x7B & 69.5 & 67.5 & 73.7 & 69.9 & 71.6 & 70.1 \\ \bottomrule
\end{tabular}
\caption{\label{table:models-results-types}
Model performance ($\%$) across the different types of questions.}
\end{table*}

Table~\ref{table:models-results-types} displays the performance of all LLMs evaluated on the benchmark across the different question types. The models do not exhibit striking differences between types, though there is a tendency toward higher results in multiple selection and matching questions. We expected these question types to be more challenging than multiple-choice variants, as they require more than simply identifying the most likely correct option. Additionally, they are evaluated using the $F_1$ score, which equally rewards correct answers and penalizes incorrect ones. Conversely, true/false questions tend to be associated with the lowest scores, which is also surprising, as we expected this type to be the easiest. Differences in the distribution of question types within each field of law may have contributed to the observed variations in performance. A deeper analysis is left for future work.

\end{document}